\documentclass[pdflatex,sn-mathphys-num]{sn-jnl}

\usepackage{graphicx}%
\usepackage{multirow}%
\usepackage{amsmath,amssymb,amsfonts}%
\usepackage{amsthm}%
\usepackage{mathrsfs}%
\usepackage[title]{appendix}%
\usepackage[dvipsnames]{xcolor}%
\usepackage{textcomp}%
\usepackage{manyfoot}%
\usepackage{booktabs}%
\usepackage{algorithm}%
\usepackage{algorithmicx}%
\usepackage{algpseudocode}%
\usepackage{subcaption}
\usepackage{listings}%
\usepackage{tikz}
\usetikzlibrary{arrows.meta, backgrounds, positioning}
\usepackage{longtable}
\usepackage{makecell}
\usepackage{array}
\usepackage{url}
\usepackage{caption}
\usepackage{float}
\DeclareMathAlphabet{\pazocal}{OMS}{zplm}{m}{n}

\theoremstyle{thmstyleone}

\theoremstyle{thmstyletwo}

\theoremstyle{thmstylethree}

\begin{document}

\newcommand{\Xp}{\pazocal{X}}
\newcommand{\Ip}{\pazocal{I}}

\title[Composite Silhouette]{Composite Silhouette: A Subsampling-based Aggregation Strategy}

\author*[1,2]{\fnm{Aggelos} \sur{Semoglou}}\email{a.semoglou@athenarc.gr}

\author[3]{\fnm{Aristidis} \sur{Likas}}\email{arly@cs.uoi.gr}

\author[1,2]{\fnm{John} \sur{Pavlopoulos}}\email{annis@aueb.gr}

\affil*[1]{
\orgname{Athens University of Economics and Business}, \orgaddress{\country{Greece}}}

\affil[2]{\orgname{Archimedes, Athena Research Center, Greece}}

\affil[3]{
\orgname{University of Ioannina}, \orgaddress{\country{Greece}}}

\abstract{Determining the number of clusters is a central challenge in unsupervised learning, where ground-truth labels are unavailable. The Silhouette coefficient is a widely used internal validation metric for this task, yet its standard micro-averaged form tends to favor larger clusters under size imbalance. Macro-averaging mitigates this bias by weighting clusters equally, but may overemphasize noise from under-represented groups. We introduce Composite Silhouette, an internal criterion for cluster-count selection that aggregates evidence across repeated subsampled clusterings rather than relying on a single partition. For each subsample, micro- and macro-averaged Silhouette scores are combined through an adaptive convex weight determined by their normalized discrepancy and smoothed by a bounded nonlinearity; the final score is then obtained by averaging these subsample-level composites. We establish key properties of the criterion and derive finite-sample concentration guarantees for its subsampling estimate. Experiments on synthetic and real-world datasets show that Composite Silhouette effectively reconciles the strengths of micro- and macro-averaging, yielding more accurate recovery of the ground-truth number of clusters.}

\maketitle

\section{Introduction}\label{sec:intro}
Clustering evaluation remains a crucial but challenging step in unsupervised learning, largely due to the inherent difficulty of assessing solutions without labeled data~\cite{jain1999data}. Among various internal clustering evaluation metrics, the Silhouette coefficient~\cite{rousseeuw1987silhouettes} is one of the most widely adopted, as it effectively quantifies the compactness and separation of clusters using only intrinsic dataset properties. Its ease of interpretation and widespread implementation in libraries such as \texttt{scikit-learn}~\cite{scikit-learn} have made it a default choice in many practical applications.
However, the traditional approach, micro-averaging Silhouette scores across all individual data points~\cite{batool2021clustering}, is known to exhibit significant bias, particularly in datasets with imbalanced cluster sizes, frequently encountered in real-world applications. In such cases, larger clusters can disproportionately influence the final score, potentially masking poor separation in minor clusters~\cite{revisiting}. 
An alternative yet rarely utilized strategy, macro-averaging, aggregates Silhouette scores by first computing the mean Silhouette per cluster and then averaging across clusters. Although macro-averaging mitigates cluster-size bias by assigning equal weight to clusters irrespective of their sizes, it can overly emphasize smaller or under-represented clusters. 
\vspace{-3mm}
\begin{figure}[htbp]
    \centering
    \begin{tikzpicture}[
        >=stealth, 
        font=\sffamily,
        header/.style={font=\sffamily\bfseries\normalsize, text=black!90},
        major/.style={circle, draw=nBlue, fill=nBlue!12, thick, minimum size=2.2cm, align=center},
        minor/.style={circle, draw=nGreen, fill=nGreen!12, thick, minimum size=0.8cm, align=center},
        weightBox/.style={rectangle, draw=#1, fill=#1!5, thick, rounded corners=4pt, text width=3.6cm, minimum height=2.4cm, align=left, inner sep=8pt},
        impactBox/.style={rectangle, draw=#1!80, fill=white, thick, rounded corners=4pt, text width=3.6cm, minimum height=2.4cm, align=left, inner sep=8pt}
    ]

    \definecolor{nBlue}{HTML}{005AB5}
    \definecolor{nGreen}{HTML}{008B45}
    \definecolor{nAmber}{HTML}{D35400}

    \node[header] at (0, 4.2) {Clustered Data};

    \node[rectangle, draw=black!25, fill=black!2, thick, rounded corners=6pt, minimum width=2.8cm, minimum height=5.6cm] (dataBox) at (0, 0) {};

    \node[major] (c1) at (-0.2, 1.3) {\Large $\boldsymbol{C_1}$};
    \node[nBlue, font=\sffamily\scriptsize] at (-0.1, 0.0) {Majority Group};

    \node[minor] (c2) at (0.4, -1.3) {\large $\boldsymbol{C_2}$};
    \node[nGreen, font=\sffamily\scriptsize] at (0.2, -1.97) {Minority Group};

    \node[header] at (4.6, 4.2) {Silhouette Aggregation};
    
    \node[weightBox=nBlue] (wMicro) at (4.6, 1.8) {
        \textbf{\normalsize Micro-Averaging} \\[4pt]
        \small \textit{Point-wise Weighting} \\[4pt]
        \small Evaluation across all data points: Every individual data point exerts equal influence on the score.
    };

    \node[weightBox=nGreen] (wMacro) at (4.6, -1.8) {
        \textbf{\normalsize Macro-Averaging} \\[4pt]
        \small \textit{Cluster-wise Weighting} \\[4pt]
        \small Evaluation across all identified clusters: Each cluster is weighted equally on the score.
    };

    \draw[->, thick, black!70] (dataBox.east) -- (wMicro.west);
    \draw[->, thick, black!70] (dataBox.east) -- (wMacro.west);

    \node[header] at (9.2, 4.2) {Domain Impact};

    \node[impactBox=nBlue] (iMicro) at (9.2, 1.8) {
        \textbf{\normalsize Business Analytics} \\[4pt]
        \small Focus on \textbf{global utility}. System performance is dominated  by the bulk population ($C_1$); minority clusters are marginal.
    };

    \node[impactBox=nGreen] (iMacro) at (9.2, -1.8) {
        \textbf{\normalsize Medical Diagnostics} \\[4pt]
        \small Focus on \textbf{patient-pattern discovery}. Minority clusters ($C_2$) represent meaningful subgroups that must remain visible.
    };

    \draw[->, thick, nBlue!90] (wMicro.east) -- (iMicro.west);
    \draw[->, thick, nGreen!90] (wMacro.east) -- (iMacro.west);

    \node[draw=nAmber, fill=nAmber!5, thick, rounded corners=4pt, text width=11.5cm, align=center, inner sep=10pt] (topic) at (4.95, -5) {
        \textbf{\large The Inductive Bias Conflict} \\[4pt]
        \normalsize It is often unclear whether a local structure represents a valuable subgroup (requiring Macro) or irrelevant statistical noise (requiring Micro).\\ Practitioners face a \textbf{bias trade-off} with no \textbf{universal selection criterion}.
    };

    \end{tikzpicture}
    \caption{Silhouette aggregation strategies: Global trends vs. local patterns.}
    \label{fig:silhouette_clean_final}
\end{figure}
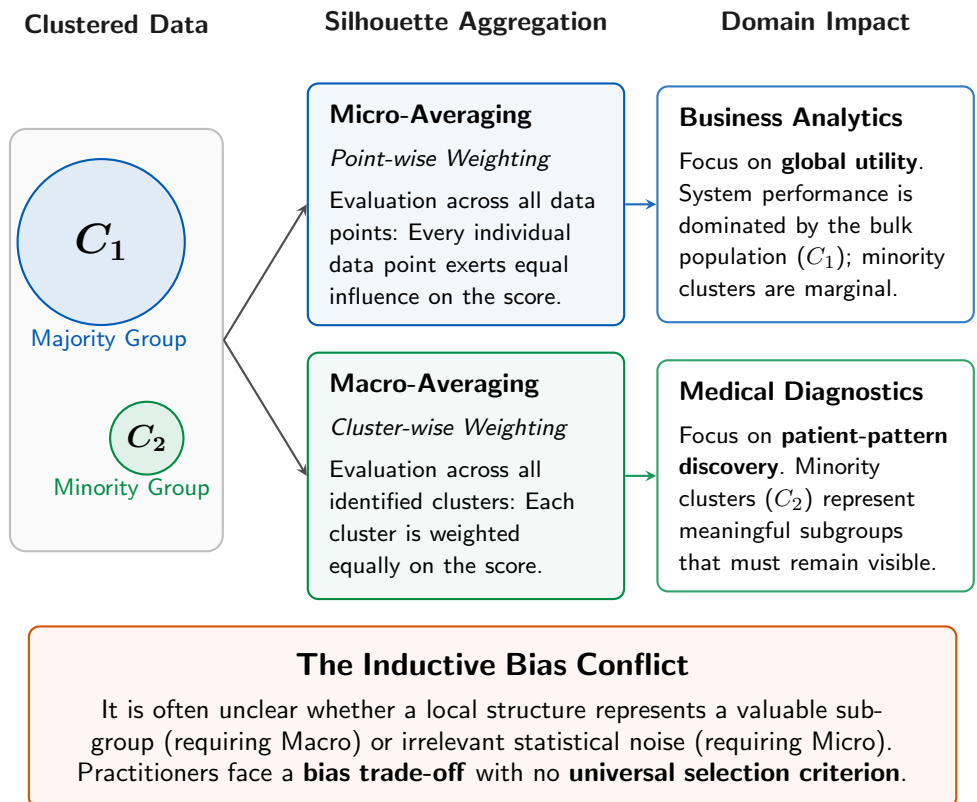
\vspace{-5mm}
This trade-off is especially evident in real-world settings. For instance, in medical clustering~\cite{med}, where rare but clinically important patient subgroups must be identified, macro-averaging is often preferable because it ensures that these small clusters are not overshadowed. In contrast, in customer segmentation~\cite{csgm}, where the business impact of large, well-defined customer groups dominates, micro-averaging tends to better reflect practical utility. Yet in other domains, such as topic clustering~\cite{topc}, where some themes are broad and frequent while others are niche but still valuable, it is often unclear which strategy should be preferred. Depending on the application, both majority coherence and minority representation may be important. Consequently, practitioners face a critical choice between two fundamentally different aggregation strategies, each having its own limitations, and with no clear rule for balancing them across datasets and clustering solutions (Fig.~\ref{fig:silhouette_clean_final}). Moreover, the disagreement between micro- and macro-averaging may itself carry useful information, as it can indicate whether clustering quality is driven primarily by majority structure or by the fidelity of smaller groups. This suggests that, rather than selecting one aggregation strategy globally, a more effective criterion should adapt to how these two perspectives agree or diverge across different views of the data. The absence of such a criterion is particularly problematic in practical settings, where internal validation metrics guide the choice of the optimal number of clusters. In such cases, inconsistent or biased metrics can lead to misleading evaluations and suboptimal clustering decisions. 
This raises a central question:

\medskip

\noindent \textit{Could an internal validation metric combine the strengths of micro- and macro-averaged Silhouette scores to guide cluster-count selection more accurately and reliably, without prior assumptions about which aggregation strategy is preferable?}

\medskip

\noindent To address this question, we introduce Composite Silhouette, a subsampling-based internal validation criterion that aggregates micro- and macro-averaged Silhouette scores over repeated subsampled clusterings. Instead of relying on a single partition or enforcing a fixed global preference between micro- and macro-averaging, our method computes the two scores on each subsampled clustering and combines them through a subsample-specific convex combination, whose mixing weight is determined by the normalized discrepancy between them and smoothed by a bounded nonlinearity. The final score is then obtained by averaging these per-subsample composite scores across subsampling trials, yielding a more stable and adaptive criterion for cluster-count selection. In this sense, Composite Silhouette performs discrepancy-driven adaptive aggregation: the relative weight assigned to the two Silhouette views is determined locally at the subsample level, rather than fixed in advance for the entire dataset. This allows the criterion to respond to heterogeneous structure without committing a priori to either majority-dominated or minority-sensitive evaluation.
We show that this formulation enjoys useful deterministic properties and admits finite-sample concentration guarantees for its subsampling estimate. Across synthetic and real-world datasets, Composite Silhouette accurately identifies the number of clusters in regimes where either micro- or macro-averaging alone would be effective, without requiring prior knowledge of which aggregation strategy is better suited to the data, while remaining competitive in balanced settings. Against a broad range of internal validation baselines, including both averages over subsampled clustering runs and averages from repeated clustering runs on the full dataset, Composite Silhouette achieves the strongest overall performance in recovering the ground-truth number of clusters.
Overall, this work offers a principled and practical alternative to existing internal validation metrics, enabling more reliable model selection in settings where traditional approaches may fail to capture the underlying cluster structure accurately. Our code for Composite Silhouette is publicly available at: \url{https://github.com/semoglou/comp_sil}.

\section{Related Work}\label{subsec:related}
\textbf{Internal cluster validity indices}

\smallskip

\noindent Internal validity indices provide a label-free way to evaluate clustering solutions by examining only the geometry and dispersion of the data~\cite{halkidi2002part1,halkidi2002part2,arbelaitz2013comparative}. Some indices focus on cluster separation, such as the Dunn index~\cite{dunn1973fuzzy}, which takes the ratio of the minimum inter-cluster distance to the maximum intra-cluster diameter to ensure worst-case separation. Other measures, like the Calinski–Harabasz criterion~\cite{calinski1974dendrite}, compare between-cluster dispersion against within-cluster dispersion, favoring compact, well-separated groups. The Davies–Bouldin index~\cite{davies1979cluster} quantifies cluster similarity by averaging, for each cluster, the worst ratio of its scatter to its nearest neighbor’s separation and selects the clustering that minimizes this average. 

\medskip

\noindent Among these, the Silhouette Coefficient stands out for its interpretability: it assigns each point a score in $[-1, 1]$ that balances cohesion against separation. Aggregated as an overall metric or examined at the per-sample level, Silhouette offers both global validation and a diagnostic for potentially misassigned points, making it a versatile tool in practice~\cite{dudek2020silhouette}. Yet, despite its popularity, the Silhouette metric, like other internal indices, can be sensitive to dataset characteristics such as cluster shape, density variation, and especially cluster-size imbalance, which may distort aggregated scores and lead to suboptimal model selection~\cite{vendramin2010relative,hassan2024review}; similar challenges also arise for unsupervised cluster-count selection procedures such as the Elbow method~\cite{thorndike1953belongs,shi2021elbow} and the Gap Statistic~\cite{tibshirani2001gap}.

\medskip

\noindent\textbf{Silhouette aggregation strategies}

\smallskip

\noindent  Most implementations of the Silhouette score rely solely on micro-averaging~\cite{shahapure2020cluster}, which aggregates the Silhouette values of all data points, effectively giving greater influence to larger clusters. While this approach is intuitive and widely used, it can introduce bias when cluster sizes are highly imbalanced. Macro-averaging, which first computes the mean Silhouette per cluster and then averages across clusters, has been proposed as an alternative to mitigate this effect~\cite{revisiting}. However, macro-averaging can overemphasize small or under-represented clusters, reducing the influence of majority groups. These two aggregation strategies therefore reflect different perspectives on clustering quality, and neither is uniformly preferable across datasets.

\medskip

\noindent \textbf{Sampling approaches and heuristics}

\smallskip

\noindent Sampling-based strategies have also been explored as a way to stabilize internal validation or reduce the computational burden of repeated clustering evaluation. Uniform subsampling can make large-scale validation more tractable, while repeated runs and averaging can reduce sensitivity to random initialization or sampling variability~\cite{lange2004stability}. In the context of Silhouette-based evaluation, such approaches may partially alleviate the dominance of large clusters or improve robustness, but they do not by themselves resolve the underlying tension between micro- and macro-aggregation. 

\medskip

\noindent More generally, heuristic combinations of validation criteria or averaging schemes have been considered, though such approaches often rely on fixed design choices or manually selected weights that may not adapt well to the structure of a given dataset~\cite{liu2010understanding}.

\medskip

\noindent\textbf{Filling the gap}

\smallskip

\noindent Although a wide range of internal validation methods has been proposed, existing approaches do not directly address how micro- and macro-averaged Silhouette scores should be combined in a data-adaptive way for cluster-count selection. In particular, current methods typically rely on a single aggregation strategy, fixed averaging rules, or standalone internal indices, without exploiting how the discrepancy between micro- and macro-level evaluations evolves across repeated subsampled clusterings. This leaves a methodological gap in settings where cluster-size imbalance or heterogeneous structure makes either aggregation strategy alone insufficient. Our work addresses this gap by introducing a subsampling-based Composite Silhouette criterion that adaptively combines micro- and macro-averaged Silhouette scores at the subsample level through a discrepancy-sensitive weighting mechanism. The resulting score provides a principled and practical approach to internal cluster-count selection without requiring prior assumptions about which aggregation perspective is more appropriate.

\section{Methodology}\label{sec:meth}
In this section, we formalize Composite Silhouette, denoted by $S_\mathrm{mM}$, and introduce the notation used throughout the method. The proposed formulation evaluates clustering quality through repeated subsampled clusterings, allowing the relationship between micro- and macro-averaged Silhouette to be assessed across multiple views of the data rather than through a single partition. The resulting construction yields an adaptive composite criterion that captures the relative behavior of the two aggregation strategies while producing a single score for cluster-count selection.

\subsection{Setup and Notation}\label{subsec:setup}
Let $\Xp=\{x_i\}_{i\in\Ip}\subset\mathbb{R}^d$ be a dataset, where $\Ip=\{1,2,\dots,N\}$ indexes the $N$ observations. Our goal is to evaluate candidate numbers of clusters by examining how clustering quality behaves across repeated subsampled views of the data, rather than relying on a single partition of the full dataset. To this end, let $\mathcal{K}\subset\mathbb{N}_{>1}$ denote the set of candidate cluster counts, and fix $k\in\mathcal{K}$. We also fix a subsampling fraction $\phi\in(0,1]$, define the subsample size as $m=\lfloor \phi N\rfloor$, and let $B\in\mathbb{N}$ denote the number of subsamples.
For each $b\in\{1,\dots,B\}$, we draw independently a subset $I_b\subset\Ip$ of size $m$, uniformly at random without replacement from $\Ip$, and define the subsample by:
\begin{equation}\label{eq:subsample}
X^{(b)}=\{x_i\in\Xp:\ i\in I_b\}\subset\Xp.
\end{equation}
These subsamples provide multiple partial views of the dataset, allowing us to study how internal validation behaves under repeated perturbations of the data. Let $F(\cdot \ ; \ k)$ denote a clustering algorithm that, given a dataset and a candidate number of clusters $k$, returns a partition into $k$ distinct (non-empty) clusters. Applied to the subsample $X^{(b)}$, it yields the partition:
\begin{equation}
    F(X^{(b)};k)=\{C_1^{(b)},\dots,C_k^{(b)}\},
\end{equation}
where $\{C_r^{(b)}\}_{r=1}^k$ are the clusters obtained on the $b$-th subsample. In this way, each candidate $k$ is associated with $B$ subsampled clusterings, which provide repeated local views of the structure induced at that value of $k$.

\subsection{Subsample Silhouette Scores}\label{subsec:silhouette}

For a fixed number of clusters $k\in\mathcal{K}$ and a fixed subsample $X^{(b)}$, we now define the Silhouette quantities associated with the partition $F(X^{(b)};k)$. For each observation $x_i\in X^{(b)}$, let $C_i^{(b)}$ denote the cluster to which $x_i$ is assigned. The Silhouette value of $x_i$ is based on two quantities: its average dissimilarity to the points in its own cluster $a_i=a_i^{(b)}(k)$, and its average dissimilarity to the nearest competing cluster $b_i=b_i^{(b)}(k)$:

\begin{align}\label{eq:sil_components}
a_i = \frac{1}{|C_i^{(b)}|-1}
\sum_{\substack{x_j\in C_i^{(b)} \\ j\neq i}}
\|x_i-x_j\|, \quad
b_i = \min_{\ell\neq c_i^{(b)}}
\frac{1}{|C_\ell^{(b)}|}
\sum_{x_j\in C_\ell^{(b)}} \|x_i-x_j\|
\end{align}
Here, $a_i$ measures how well $x_i$ is embedded within its assigned cluster, while $b_i$ measures its average dissimilarity to the closest alternative cluster. Smaller values of $a_i$ indicate stronger within-cluster cohesion, whereas larger values of $b_i$ indicate stronger separation from neighboring clusters.
Using these two quantities, the Silhouette score of $x_i$, $s_i = s_i(k)$, on the $b$-th subsample is given by:
\begin{equation}\label{eq:s_subsample}
s_i^{(b)}=\frac{b_i-a_i}{\max\{a_i,b_i\}}\in[-1,1].
\end{equation}
A value close to $1$ indicates that the observation is well matched to its assigned cluster and well separated from competing clusters, a value near $0$ indicates ambiguity, and a negative value suggests that the observation may fit better in another cluster.
We next aggregate these per-instance values in two different ways: micro-averaged ($S_\mathrm{m}^{(b)}$) and macro-averaged ($S_\mathrm{M}^{(b)}$) Silhouette scores:
\begin{equation}\label{eq:micro_macro_sil}
S_\mathrm{m}^{(b)}=\frac{1}{m}\sum_{i\in I_b} s_i^{(b)}, \quad S_\mathrm{M}^{(b)}=
\frac{1}{k}\sum_{r=1}^{k}
\frac{1}{|C_r^{(b)}|}
\sum_{x_i\in C_r^{(b)}} s_i^{(b)}.
\end{equation}
These correspond to two complementary ways of evaluating clustering quality on the same subsample. The micro-averaged score ($S_\mathrm{m} \in [-1,1]$) assigns equal weight to all observations and is therefore more strongly influenced by the structure of larger clusters. In contrast, the macro-averaged score ($S_\mathrm{M} \in [-1,1]$) first averages within clusters and then across clusters, so that each cluster contributes equally regardless of its size. As a result, it is more sensitive to the quality of smaller groups.
These scores form the components of our composite criterion. In the next subsection, we compare these subsample-level quantities and use their relationship to define the weighting scheme underlying Composite Silhouette.

\subsection{Subsample-Level Discrepancy and Weighting Scheme}\label{subsec:weighting}

For a fixed candidate number of clusters $k$, $S_\mathrm{m}^{(b)}$ and $S_\mathrm{M}^{(b)}$ may evaluate the same subsampled clustering differently. We quantify this difference on the $b$-th subsampled clustering through the discrepancy:
\begin{equation}\label{eq:db}
    \Delta_b = S_\mathrm{m}^{(b)} - S_\mathrm{M}^{(b)}.
\end{equation}
When $\Delta_b>0$, micro-averaging assigns a higher score than macro-averaging, indicating that the clustering quality on that subsample is reflected to a greater extent by the observation-level, global view. Conversely, when $\Delta_b<0$, macro-averaging assigns the higher score, indicating that the evaluation is better represented by the cluster-level view. Thus, $\Delta_b$ captures both the direction and the magnitude of the disagreement between the two aggregation strategies.
This discrepancy is not meant to declare one aggregation inherently correct. Rather, it serves as a local signal of how the clustering structure revealed by a given subsample is reflected by the two Silhouette aggregations. Instead of imposing a fixed preference for either perspective, we use this local disagreement to balance micro- and macro-averaging adaptively within each subsampled clustering.
Since the magnitude of $\Delta_b$ may vary across $b \in \{1,\dots,B\}$, we normalize it relative to the largest absolute discrepancy observed $\Delta_{\max} = \max_{1\leq b \leq B} |\Delta_b|$: 
\begin{equation}\label{eq:norm_db}
\widetilde{\Delta}_b = \frac{\Delta_b}{\Delta_{\max}+\varepsilon} \in (-1,1), \quad \varepsilon>0 \text{ small for numerical stability}.
\end{equation}
This ensures that the difference is interpreted relative to the range of disagreement observed across the current collection of subsampled clusterings, rather than through its raw scale.
To obtain a smooth and stable transformation of the normalized difference, we pass $\widetilde{\Delta}_b$ through the hyperbolic tangent function: 
\begin{equation}\label{eq:zb}
z_b = \tanh(\widetilde{\Delta}_b).
\end{equation}
This transformation is monotone and nearly linear around zero, so small values of $\widetilde \Delta_b$ are preserved almost proportionally, whereas larger ones are gradually compressed. Its smoothness, symmetry, and gradual saturation make $\tanh(\cdot)$, a natural choice for ``encoding'' local micro--macro disagreement in a stable and interpretable way (see Appendix~\ref{app:sec:ablation} Table~\ref{app:tab:abl} for an ablation over alternative transformations, including linear, nonlinear, and hard-threshold mappings).

\noindent For each subsample $b \in \{1, \dots,B\}$, we use $z_b$ to combine $S_\mathrm{m}^{(b)}$ and $S_\mathrm{M}^{(b)}$ in a way that reflects their relative behavior on that subsample;
we map $z_b \in (-1,1)$ to $(0,1)$ and define the subsample-specific convex weight assigned to $S_\mathrm{m}^{(b)}$ as:
\begin{equation}\label{eq:wb}
w_b = \frac{1+z_b}{2} \in (0,1),
\end{equation}
so that the weight assigned to $S_\mathrm{M}^{(b)}$ is $1-w_b$. By construction the weighting remains centered around $\tfrac{1}{2}$. In particular, when $\Delta_b >0$, we have $w_b > \frac{1}{2}$, so the combination places greater emphasis on micro-averaging; when $\Delta_b < 0$, then $w_b<\frac{1}{2}$, so the emphasis shifts toward macro-averaging. If $\Delta_b \approx 0\Rightarrow w_b \approx \frac{1}{2}$, meaning the two strategies are weighted equally. Thus, the sign of $\Delta_b$ determines which aggregation is favored on a given subsample, while its transformed magnitude determines how strongly that preference is expressed. This yields a subsample-specific balancing mechanism between the two Silhouette aggregations, favoring the strategy that is better supported on each subsampled clustering while still accounting for the other in proportion to the magnitude of their discrepancy, namely $w_bS_\mathrm{m}^{(b)} + (1-w_b)S_\mathrm{M}^{(b)}$. 

\subsection{Composite Silhouette}\label{subsec:smm}
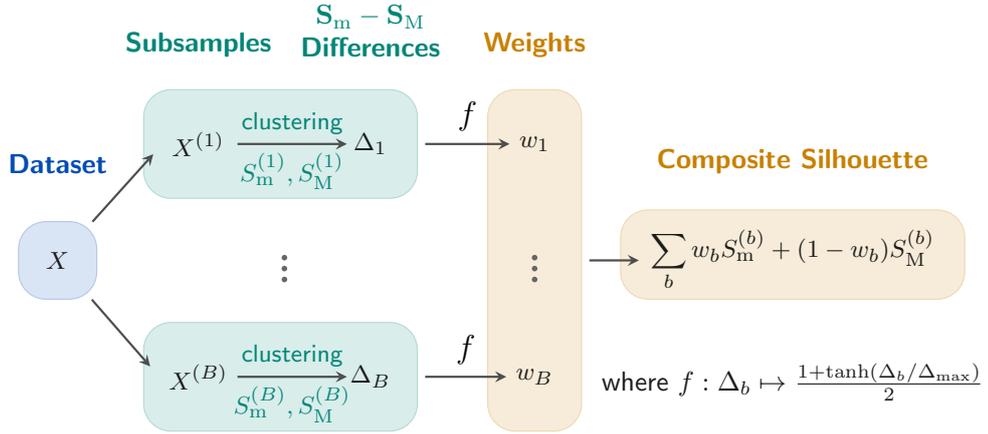
\begin{figure}[h]
    \centering
    \resizebox{\linewidth}{!}{%
    \begin{tikzpicture}[
        >=stealth,
        arrow/.style={->, thick, draw=black!70},
        fanArrow/.style={->, thick, draw=black!70, shorten <=19pt, shorten >=5pt},
        labelStyle/.style={font=\small\sffamily, align=center, text=black!90} 
    ]

    \definecolor{nBlue}{HTML}{004CB4}
    \definecolor{nGreen}{HTML}{008577}
    \definecolor{nAmber}{HTML}{C88000}

    \begin{scope}[on background layer]
        
        \filldraw[draw=nBlue!30, fill=nBlue!15, rounded corners=10pt] (-0.5,-0.5) rectangle (0.5,0.5);

        \filldraw[draw=nGreen!30, fill=nGreen!15, rounded corners=10pt] (1.1,2.2) rectangle (4.6,0.8);
        \filldraw[draw=nGreen!30, fill=nGreen!15, rounded corners=10pt] (1.1,-0.8) rectangle (4.6,-2.2);

        \filldraw[draw=nAmber!30, fill=nAmber!15, rounded corners=10pt] (5.5,2.2) rectangle (6.7,-2.2);

        \filldraw[draw=nAmber!30, fill=nAmber!15, rounded corners=10pt] (7.2,-0.5) rectangle (11.6,0.65);
    \end{scope}

    \node[anchor=south, font=\normalsize\bfseries\sffamily, text=nBlue] at (0, 1) {Dataset};
    \node[anchor=south, font=\normalsize\bfseries\sffamily, text=nGreen] at (1.8, 2.5) {Subsamples};
    \node[anchor=south, align=center, font=\normalsize\bfseries\sffamily, text=nGreen] at (4.0, 2.5) {$\mathbf{S_\mathrm{m}-S_\mathrm{M}}$\\Differences};
    \node[anchor=south, font=\normalsize\bfseries\sffamily, text=nAmber] at (6.1, 2.5) {Weights}; 
    \node[anchor=south, font=\normalsize\bfseries\sffamily, text=nAmber] at (9.4, 1) {Composite Silhouette};

    \node[font=\small\bfseries, text=black!90] (data) at (0, 0) {$X$};

    \node[font=\small\bfseries, text=black!90] (sub1) at (1.8, 1.5) {$X^{(1)}$};
    \node[font=\small, text=black!90] (diff1) at (4.0, 1.5) {$\Delta_1$};
    \node[font=\small\bfseries, text=black!90] (w1) at (6.1, 1.5) {$w_1$};

    \draw[fanArrow] (data.center) -- (sub1.west); 
    
    \draw[arrow] (2.3, 1.5) -- node[labelStyle, above, text=nGreen] {clustering}
                               node[labelStyle, below, inner sep=3pt, text=nGreen] {$S_\mathrm{m}^{(1)}, S_\mathrm{M}^{(1)}$} (3.7, 1.5);
    
    \draw[arrow] (4.7, 1.5) -- node[above, yshift=1pt, font=\large\itshape] {$f$} (w1.west);

    \node[text=black!60] at (2.9, 0) {\Huge $\vdots$};
    \node[text=black!60] at (6.1, 0) {\Huge $\vdots$};

    \node[font=\small\bfseries, text=black!90] (subB) at (1.8, -1.5) {$X^{(B)}$};
    \node[font=\small, text=black!90] (diffB) at (4.0, -1.5) {$\Delta_B$};
    \node[font=\small\bfseries, text=black!90] (wB) at (6.1, -1.5) {$w_B$};

    \draw[fanArrow] (data.center) -- (subB.west);

    \draw[arrow] (2.3, -1.5) -- node[labelStyle, above, text=nGreen] {clustering} 
                                node[labelStyle, below, inner sep=3pt, text=nGreen] {$S_\mathrm{m}^{(B)}, S_\mathrm{M}^{(B)}$} (3.7, -1.5);
    
    \draw[arrow] (4.7, -1.5) -- node[above, yshift=1pt, font=\large\itshape] {$f$} (wB.west);

    \node[inner sep=4pt, font=\small, text=black!90] (final) at (9.4, 0) {$\displaystyle \sum_{b} w_b S_\mathrm{m}^{(b)} + (1-w_b)S_\mathrm{M}^{(b)}$};

    \node[anchor=north, font=\normalsize\sffamily, text=black!90] at (9.4, -1.2) {where $f: \Delta_b \mapsto \frac{1 + \tanh(\Delta_b / \Delta_{\max})}{2}$};

    \draw[->, thick, draw=black!70] (6.8, 0) -- (final.west);

    \end{tikzpicture}%
    }
    \vspace{0.5em}
    \caption{Overview of the $S_{\mathrm{mM}}$ evaluation for a given number of clusters $k$.}
    \label{fig:conceptual_architecture}
\end{figure}

For a fixed candidate number of clusters $k$, the Composite Silhouette score is obtained by averaging the $B$ subsample convex combinations induced by the weights $w_b$ (Eq.~\ref{eq:wb}):
\begin{equation}\label{eq:smm}
    S_\mathrm{mM} = \frac{1}{B}\sum_{b=1}^{B}\left[w_bS_\mathrm{m}^{(b)} + (1-w_b)S_\mathrm{M}^{(b)} \right]
\end{equation}
This yields a single criterion that summarizes clustering quality across repeated subsampled views of the data while preserving how the relative support for micro- and macro-averaging varies from one subsample to another (Fig.~\ref{fig:conceptual_architecture}).

\vspace{-5mm}
\noindent \textbf{Properties}

\smallskip

\noindent For convenience, let $S_\mathrm{mM}^{(b)} := w_bS_\mathrm{m}^{(b)} + (1-w_b)S_\mathrm{M}^{(b)}$
denote the composite score on the $b$-th subsample. Since $w_b\in(0,1)$, each $S_\mathrm{mM}^{(b)}$ is a convex combination of $S_\mathrm{m}^{(b)}$ and $S_\mathrm{M}^{(b)}$, and therefore satisfies $\min\{S_\mathrm{m}^{(b)},S_\mathrm{M}^{(b)}\}
\leq
S_\mathrm{mM}^{(b)}
\leq
\max\{S_\mathrm{m}^{(b)},S_\mathrm{M}^{(b)}\}.$
Substituting $w_b=\tfrac{1+z_b}{2}$ and $\Delta_b=S_\mathrm{m}^{(b)}-S_\mathrm{M}^{(b)}$ gives (see detailed analysis in Appendix~\ref{app:subsec:prop1}):
\begin{equation}
S_\mathrm{mM}^{(b)}
=
\frac{S_\mathrm{m}^{(b)}+S_\mathrm{M}^{(b)}}{2}
+
\frac{\Delta_b z_b}{2}
=
\frac{S_\mathrm{m}^{(b)}+S_\mathrm{M}^{(b)}}{2}
+
\frac{\Delta_b}{2}
\tanh\!\left(\frac{\Delta_b}{\Delta_{\max}+\varepsilon}\right).
\end{equation}
Thus, each subsample-level composite starts from the midpoint of the micro- and macro-averaged Silhouette scores and is then pulled toward the larger one by an amount controlled by the transformed difference. Averaging over the $B$ subsamples:
\[
\frac{1}{B}\sum_{b=1}^{B}\min\{S_\mathrm{m}^{(b)},S_\mathrm{M}^{(b)}\}
\leq
S_\mathrm{mM}
\leq
\frac{1}{B}\sum_{b=1}^{B}\max\{S_\mathrm{m}^{(b)},S_\mathrm{M}^{(b)}\} \Rightarrow S_\mathrm{mM}\in[-1,1].
\]
Notably, although each $S_\mathrm{mM}^{(b)}$ lies between its corresponding micro- and macro-averaged components, the final score $S_\mathrm{mM}$ need not lie between the sample-averaged quantities
\begin{equation}\label{eq:subavg}
\overline S_\mathrm{m}=\frac{1}{B}\sum_{b=1}^B S_\mathrm{m}^{(b)},
\qquad
\overline S_\mathrm{M}=\frac{1}{B}\sum_{b=1}^B S_\mathrm{M}^{(b)},
\end{equation}
since our proposed formulation averages subsample-specific convex combinations rather than applying a single global convex combination to $\overline S_\mathrm{m}$ and $\overline S_\mathrm{M}$ (illustrated in Appendix Figs.~\ref{app:fig:trends1}--\ref{app:fig:trends6}). Indeed, using $w_b=\tfrac{1+z_b}{2}$ (Eq.~\ref{eq:wb}) and $\Delta_b=S_\mathrm{m}^{(b)}-S_\mathrm{M}^{(b)}$ (Eq.~\ref{eq:db}), we obtain (detailed analysis in Appendix~\ref{app:subsec:prop2}):
\begin{equation}
S_\mathrm{mM}
=
\frac{\overline S_\mathrm{m}+\overline S_\mathrm{M}}{2}
+
\frac{1}{2B}\sum_{b=1}^B \Delta_b z_b.
\end{equation}
Thus, $S_\mathrm{mM}$ is the midpoint of $\overline S_\mathrm{m}$ and $\overline S_\mathrm{M}$ plus a discrepancy-dependent correction term, rather than a single convex combination of the two sample-averaged scores.

\medskip

\noindent \textbf{Cluster count selection with } $\mathbf{S_\mathrm{mM}}$

\smallskip

\noindent Evaluating $S_\mathrm{mM}$ over the candidate set $\mathcal{K}$, allows us to select the number of clusters as the value of $k$ that maximizes the Composite Silhouette score:\footnote{A more conservative alternative (supported by our implementation) is to select $k$ by maximizing a lower confidence bound (LCB) of $S_\mathrm{mM}(k)$, thus favoring candidates with both high score and low subsampling variability.}
\begin{equation}\label{eq:max}
    k^\star = \arg \max_{k \in \mathcal{K}}S_\mathrm{mM}(k).
\end{equation}

\noindent \textbf{Complexity}

\smallskip

\noindent Let $T_F(m,d,k)$ denote the cost of applying the clustering algorithm $F(\cdot;k)$ to a subsample of size $m$ in dimension $d$. For a fixed candidate number of clusters $k$, clustering across the $B$ subsamples costs $\mathcal{O}\!\left(B\,T_F(m,d,k)\right)$, while computing the corresponding subsample-level micro- and macro-averaged Silhouette scores costs $\mathcal{O}(B\,m^2d)$. The remaining operations specific to $S_\mathrm{mM}$---namely the computation of differences, transformations, weights, and the final averaging---are linear in $B$ and therefore negligible. Hence, for a fixed $k$, the overall complexity is $\mathcal{O}\!\left(B\,T_F(m,d,k)+B\,m^2d\right)$.

\medskip

\noindent In the case of $k$-means: $T_F(m,d,k)=\mathcal{O}(kmd)$, yielding $\mathcal{O}(Bkmd+Bm^2d)$. Thus, $S_\mathrm{mM}$ preserves the dominant computational profile of repeated subsampled Silhouette evaluation while adding only negligible discrepancy-aware overhead.

\section{Theoretical Analysis}\label{sec:theoretical}

We briefly summarize the main probabilistic guarantee underlying Composite Silhouette; full statements, extensions, and proofs are deferred in Appendix~\ref{app:sec:theory}. For the analysis, we consider the candidate set $\mathcal K$ and assume that the subsamples are drawn independently according to the sampling scheme, while any randomness of the clustering algorithm is also independent across trials and independent of the subsampling. In our method, the normalizing quantity $\Delta_{\max} = \Delta_{\max}(k)$ is computed from the same subsamples used to form $S_\mathrm{mM}$, which induces dependence across the resulting terms. To obtain concentration bounds with standard tools, we therefore analyze a closely related version in which $\Delta_{\max}$ is estimated from an independent set of subsamples and then used to form the final score. This preserves the form of the method while ensuring that the per-subsample composite scores used in the analysis are independent and identically distributed. In practice, this distinction is mainly technical: the independent estimation of $\Delta_{\max}$ is introduced to obtain clean concentration bounds, while the empirical behavior remains very similar to that of the implemented version for moderate to large values of $B$.
Under this setup, each subsample composite $w_bS_\mathrm{m}^{(b)} + (1-w_b)S_\mathrm{M}^{(b)}$ lies in $[-1,1]$, so Hoeffding's inequality applies directly. In particular, when the candidate set $\mathcal K$ is finite, for any $\delta\in(0,1)$, the estimate
$\sup_{k\in\mathcal K}\left|S_\mathrm{mM}(k)-\mathbb E\!\left[S_\mathrm{mM}^{(b)}(k)\mid \Delta_{\max}(k)\right]\right|
\le
\sqrt{\frac{2\ln(2|\mathcal K|/\delta)}{B}}$
holds with conditional probability at least $1-\delta$ (proof in Appendix~\ref{app:subsec:uniform}).
Thus, the entire curve $k\mapsto S_\mathrm{mM}(k)$ is uniformly estimated at rate $O(B^{-1/2})$, so the relative ordering of candidate cluster counts is increasingly stable as the number of subsamples grows. The corresponding fixed-$k$ concentration bound is given in Appendix~\ref{app:subsec:fixedk}, while Appendix~\ref{app:subsec:recovery} shows that, under a standard positive-margin condition on the subsampling objective (i.e., the optimal candidate is separated from all others by a strictly positive gap in the subsampling expectation), the probability of selecting the optimal candidate approaches one as $B$ increases. Together, these results show that increasing the number of subsamples improves both the accuracy of the estimated Composite Silhouette curve and the reliability of the resulting choice of $k^\star$.

\section{Empirical Validation}\label{sec:empirical}

\subsection{Synthetic and Real-World Datasets}\label{subsec:data} 
We evaluate Composite Silhouette on four synthetic datasets and twelve real-world datasets spanning balanced and imbalanced cluster structures, heterogeneous cluster scales, tabular data, image representations, and text embeddings.  

\medskip

\begin{figure}[H]
    \centering

    \begin{subfigure}[t]{0.45\textwidth}
        \centering
        \includegraphics[width=\textwidth]{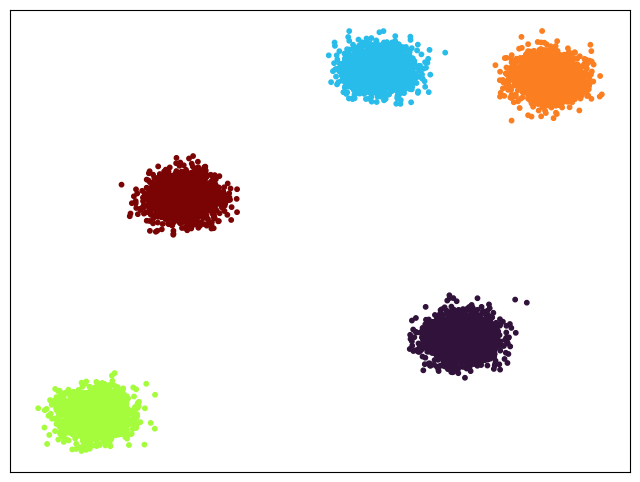}
        \caption{S1}
        \label{fig:s1}
    \end{subfigure}
    \hfill
    \begin{subfigure}[t]{0.45\textwidth}
        \centering
        \includegraphics[width=\textwidth]{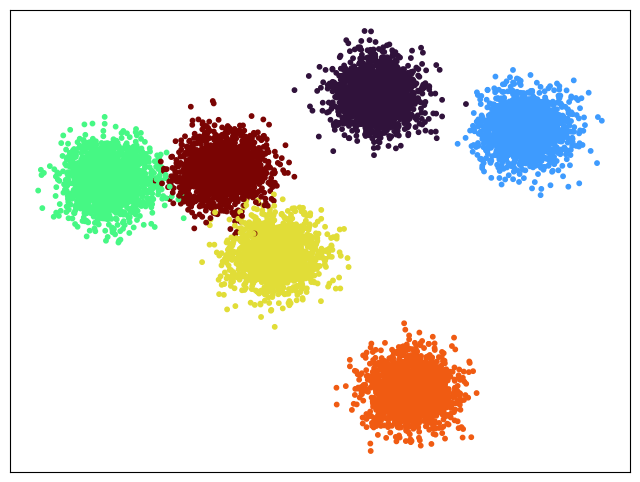}
        \caption{S2}
        \label{fig:s2}
    \end{subfigure}

    \vspace{0.8em}

    \begin{subfigure}[t]{0.45\textwidth}
        \centering
        \includegraphics[width=\textwidth]{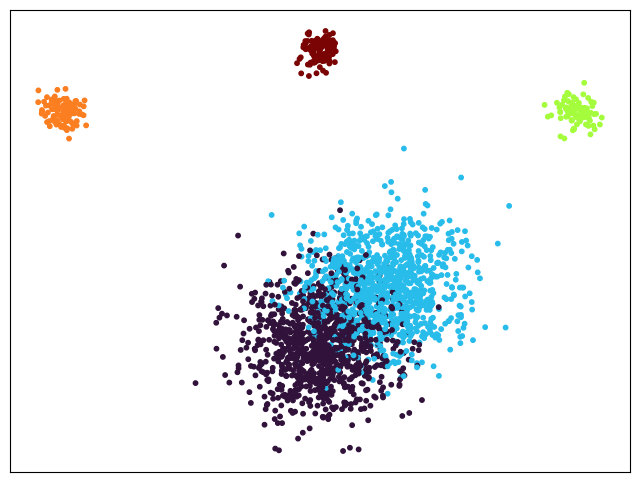}
        \caption{S3}
        \label{fig:s3}
    \end{subfigure}
    \hfill
    \begin{subfigure}[t]{0.45\textwidth}
        \centering
        \includegraphics[width=\textwidth]{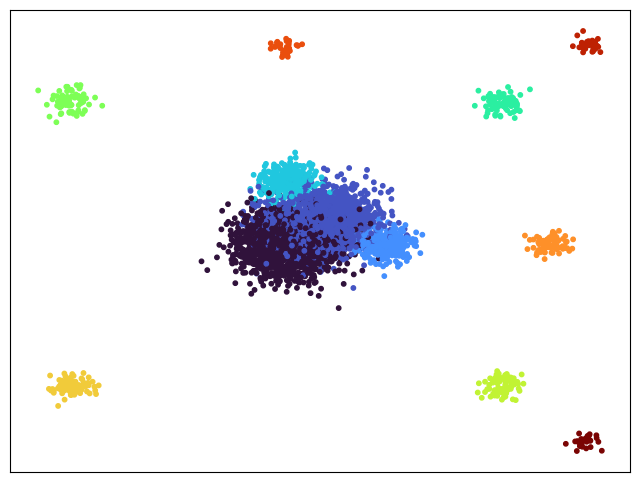}
        \caption{S4}
        \label{fig:s4}
    \end{subfigure}

    \caption{Synthetic datasets (\textbf{S1}-\textbf{S4}), colored by ground-truth cluster labels.}
    \label{fig:synthetic_datasets}
\end{figure}

\noindent The \textbf{synthetic datasets} (S1--S4, visualization in Fig.~\ref{fig:synthetic_datasets}) are designed to probe different structural regimes, from clean and well-separated clusters to strongly imbalanced and heterogeneous mixtures:
\noindent\textbf{S1} consists of 10,000 samples generated from 5 Gaussian clusters with standard deviation $0.5$. This dataset provides a clean and nearly balanced benchmark with sharply separated groups, for which we expect both aggregation strategies to perform well (Eq.~\ref{eq:micro_macro_sil}).
\textbf{S2} consists of 10,000 samples generated from 6 Gaussian clusters with standard deviation $0.9$. Relative to S1, the clusters are less sharply separated, yielding a balanced but more diffuse setting in which micro-averaging is expected to provide the stronger signal.
\textbf{S3} contains 2,300 samples arranged in 5 clusters with pronounced size and scale imbalance: two large clusters of 1,000 samples each and three small clusters of 100 samples each. The two large clusters are placed relatively close to one another and have larger spread ($\sigma=1.8$), whereas the three small clusters are more compact ($\sigma=0.5$). This dataset is intended to highlight the tension between observation-level and cluster-level aggregation under imbalance, and we expect macro-averaging to be more informative in this setting.
\textbf{S4} contains 4,090 samples distributed over 12 clusters with multiple size regimes: two large clusters (1,500 samples each), two medium clusters (300 each), five small clusters (80 each), and three tiny clusters (30 each). The cluster spreads vary across groups ($\sigma\in\{2.2,1.2,0.9,0.6\}$), producing a challenging setting with substantial heterogeneity in both scale and cluster size, in which we do not expect either micro- or macro-averaging alone to perform consistently well.

\noindent The \textbf{real-world datasets} include binary and multiclass data from tabular, vision, and text domains:\footnote{All real datasets are available via \textsc{scikit-learn}, OpenML, UCI, and Hugging Face Datasets.}
\noindent\textbf{Parkinsons} (\textsc{Pks}) contains $195$ voice recordings described by $22$ features and forms $2$ classes. \textbf{Wine} (\textsc{Wne}) contains $178$ wine samples with $13$ chemical attributes and $3$ classes.
\textbf{Blood Transfusion} (\textsc{Bld}) contains $748$ donor records with $4$ numerical features and $2$ classes.
\textbf{Digits} (\textsc{Dgt}) contains $1{,}797$ handwritten digit images with $64$ features and $10$ classes.
\textbf{BBC News} (\textsc{Bbc}) contains $2{,}225$ news articles across $5$ topics; documents are embedded with \texttt{all-MiniLM-L6-v2}~\cite{reimers2019-SentenceBERT} and reduced to $50$ dimensions by PCA.
\textbf{HTRU2} (\textsc{Htr}) contains $17{,}898$ astrophysical signal instances described by $8$ features and $2$ classes, with strong class imbalance.
\textbf{STL-10} (\textsc{Stl}) contains $5{,}000$ natural images from $10$ classes; images are represented by pretrained ResNet-18 embeddings of dimension $512$.
\textbf{20 Newsgroups} (\textsc{Nsg}) contains $11{,}314$ training documents from $20$ topic categories; we use TF--IDF representations, reduced to $100$ dimensions via truncated SVD.
\textbf{Spambase} (\textsc{Spm}) contains $4{,}601$ email instances represented by $57$ numerical features and $2$ classes.
\textbf{Minds-14} (\textsc{Mds}) contains $2{,}797$ utterances from $14$ intent classes; utterances are embedded with \texttt{all-MiniLM-L6-v2}, and reduced to $100$ dimensions by PCA, and $\ell_2$-normalized.
\textbf{Bank Marketing} (\textsc{Bnk}) contains $45{,}211$ customer records with $7$ features and $2$ classes.
\textbf{Banking77} (\textsc{B77}) contains $13{,}083$ utterances from $77$ intent classes; utterances are embedded with \texttt{all-MiniLM-L6-v2}, reduced to $50$ dimensions by PCA, and $\ell_2$-normalized.
For tabular datasets, numerical features were standardized before clustering.

\vspace{-3mm}

\subsection{Experiments}\label{subsec:experiments}
We compare $S_\mathrm{mM}$ (Eq.~\ref{eq:smm}) against a broad set of internal validation baselines under a common \emph{k}-means evaluation protocol. The comparison includes subsample-averaged micro- and macro-averaged Silhouette scores $\overline S_\mathrm{m}, \ \overline S_\mathrm{M}$ (Eq.~\ref{eq:subavg}), computed on exactly the same subsamples used by $S_\mathrm{mM}$, as well as repeated full-data averages of micro- and macro-averaged Silhouette (avg $S_\mathrm{m}$, avg $S_\mathrm{M}$), Calinski--Harabasz (avg CH), and Davies--Bouldin (avg DB), where the number of repeated runs matches the number of subsamples used by $S_\mathrm{mM}$. Here, ``full-data average'' refers to averaging the corresponding index across repeated \emph{k}-means runs on the full dataset under different random initializations. For completeness, we also report the number of clusters selected by the mean Elbow criterion (avg EL) and by the Gap statistic (GAPs). For each dataset, we evaluate methods over a candidate set of cluster counts centered at the ground-truth value, namely $\mathcal{K}=\{k_{\mathrm{GT}}-5,\dots,k_{\mathrm{GT}}+5\}$. When $k_{\mathrm{GT}}<5$, we instead use $\mathcal{K}=\{2,\dots,k_{\mathrm{GT}}+5\}$, ensuring that the search range always begins at $2$.
For each score-based method, we report the selected number of clusters (Table~\ref{tab:main_kmeans_results}) and the value attained at the ground-truth number of clusters $k_{\mathrm{GT}}$ (\S \ref{subsec:data}, Appendix~\ref{app:sec:val}; Table~\ref{app:tab:supp_kgt_values}). Trend figures over candidate values of $k$ for $S_\mathrm{mM}$, $\overline S_\mathrm{m}$, and $\overline S_\mathrm{M}$, together with the corresponding results obtained using GMM, and Bisecting \emph{k}-means are provided in Appendix~\ref{app:sec:val}) (Figs~\ref{app:fig:trends1}--\ref{app:fig:trends6}, Tables~\ref{app:tab:bisecting_main_results}--\ref{app:tab:gmm_main_results}).

\medskip 

\noindent \textbf{$S_\mathrm{mM}$ parameters}

\smallskip

\noindent Across all experiments, we use  $B=20-30$ subsamples and the subsample size is selected automatically from the dataset size $N$ and the largest candidate number of clusters $k_{\max}$. Specifically, the subsample size is chosen as the larger of two quantities: a fraction ($\phi$) of the dataset size, and a minimum size of $30$ observations per candidate cluster at $k_{\max}$. The fraction is set to $80\%$ of the data for datasets with at most $2{,}000$ samples, $60\%$ for datasets with between $2{,}001$ and $20{,}000$ samples, and $40\%$ for larger datasets. The resulting value is then capped at the full dataset size. In this way, the rule preserves sufficient cluster representation for large candidate values of $k$ while keeping the computational cost manageable on larger datasets.
\vspace{-6mm}
\begin{table*}[h]
\centering
\small
\setlength{\tabcolsep}{3.8pt}
\caption{Selected number of clusters obtained via $S_\mathrm{mM}$, $\overline S_\mathrm{m}$, $\overline S_\mathrm{M}$, avg $S_\mathrm{m}$, avg $S_\mathrm{M}$, avg CH, avg DB, avg EL, and GAPs under the $k$-means protocol (values at $k_\mathrm{GT}$ are reported in Appendix~\ref{app:sec:val}, Table~\ref{app:tab:supp_kgt_values}). All score-based criteria select $k$ by maximizing their respective scores, except avg DB, which is minimized. \textbf{Bold} entries indicate that the selected $k$ matches the ground-truth $k_\mathrm{GT}$.}
\vspace{0.5mm}
\label{tab:main_kmeans_results}
\begin{tabular}{lccccccccc}
\toprule
Dataset & $S_\mathrm{mM}$ & $\overline S_\mathrm{m}$ & $\overline S_\mathrm{M}$ & avg $S_\mathrm{m}$ & avg $S_\mathrm{M}$ & avg CH & avg DB & avg EL & GAPs \\
\midrule
\textsc{S1}  & \textbf{5}   & \textbf{5}  & \textbf{5}   & \textbf{5}   & \textbf{5}   & \textbf{5}  & \textbf{5}  & \textbf{5} & \textbf{5} \\
\textsc{S2}  & \textbf{6}  & \textbf{6}  & 3          & \textbf{6}  & 3          & \textbf{6}   & \textbf{6}  & 3 & \textbf{6} \\
\textsc{S3}  & \textbf{5}   & 2           & \textbf{5}   & 2           & \textbf{5}   & 10           & \textbf{5}  & \textbf{5} & 3 \\
\textsc{S4}  & \textbf{12}  & 8          & 13           & 9          & 9           & 16         & 9          & 9 & 14 \\
\midrule
\textsc{Pks} & \textbf{2}   & \textbf{2}   & \textbf{2}   & \textbf{2}   & \textbf{2}  & 3             & \textbf{2}  & 3 & 7 \\
\textsc{Wne} & \textbf{3}  & \textbf{3}   & \textbf{3}   & \textbf{3}   & \textbf{3}   & \textbf{3}     & \textbf{3}  & \textbf{3} & \textbf{3} \\
\textsc{Bld} & \textbf{2}  & \textbf{2} & 6            & \textbf{2}   & 6            & 6            & 4           & 4 & 5 \\
\textsc{Dgt} & \textbf{10} & \textbf{10} & \textbf{10}  & \textbf{10} & 9       & 3            & 9           & 3 & \textbf{10} \\
\textsc{Bbc} & \textbf{5}   & 6           & \textbf{5}   & \textbf{5}  & \textbf{5}   & 2           & 7         & \textbf{5} & 9 \\
\textsc{Htr} & \textbf{2}  & \textbf{2}   & \textbf{2}  & \textbf{2}   & \textbf{2}  & 4            & 3      & 4 & 7 \\
\textsc{Stl} & \textbf{10} & \textbf{10} & \textbf{10} & 6         & 6    & 6         & \textbf{10} & 8 & 14 \\
\textsc{Nsg} & \textbf{20} & 24        & 21        & 23      & 24    & 24          & 24         & 21 & 24 \\
\textsc{Spm}  & \textbf{2}  & \textbf{2}  & \textbf{2}   & \textbf{2}  & \textbf{2}   & 3            & \textbf{2}  & 3 & 3 \\
\textsc{Mds} & \textbf{14}  & 15          & 12       & 15           & \textbf{14}  & \textbf{14}    & \textbf{14}  & \textbf{14} & 16 \\
\textsc{Bnk} & \textbf{2}   & \textbf{2}   & \textbf{2}  & \textbf{2}   & \textbf{2}   & 8            & 8         & 8 & \textbf{2} \\
\textsc{B77} & \textbf{77}  & 82           & \textbf{77}  & 81          & \textbf{77}  & 72          & 81        & 81 & 75 \\
\bottomrule
\end{tabular}
\end{table*}
\normalsize

\begin{figure}[H]
    \centering
    \includegraphics[width=\textwidth]{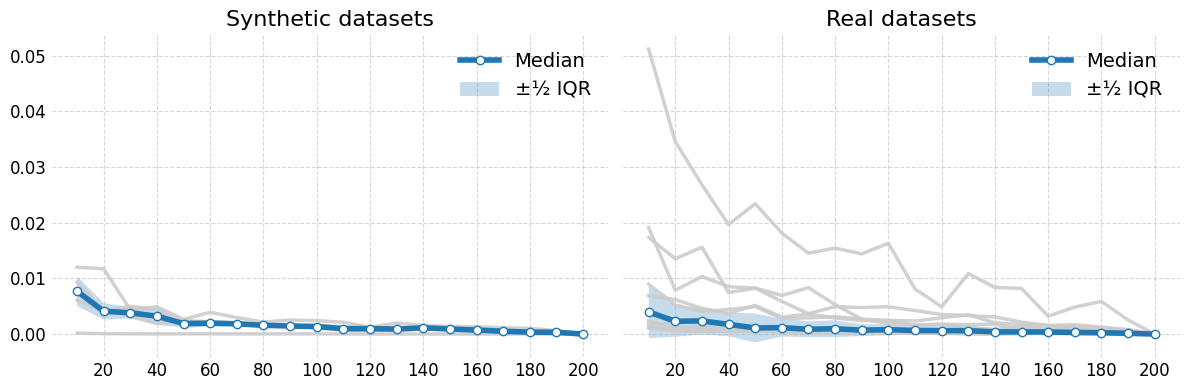}
        \caption{Approximation error of $S_\mathrm{mM}$ as a function of the number of subsamples $B$ (left: synthetic datasets, right: real-world datasets). Curves: \textcolor{gray}{gray}: per-dataset median absolute error, \textcolor{MidnightBlue}{blue}: across-dataset median, shaded band: $\pm \tfrac{1}{2}$ IQR.}
    \label{fig:num_samples}
\end{figure}

\noindent To examine how the Composite Silhouette score stabilizes as the number of subsamples $B$ increases, we study its approximation error as a function of $B$ at the ground-truth number of clusters $k_{\mathrm{GT}}$. For each dataset, we first compute a reference estimate using $B_{\max}{=}200$ subsamples under the same automatic subsample-size rule described above. We then treat this $B_{\max}$ estimate as a high-precision proxy and, for each smaller value $B\in\{10,20,\dots,200\}$, approximate it by recomputing $S_\mathrm{mM}$ from only $B$ subsamples drawn without replacement from the full pool of 200. This subsampling of subsamples is repeated $25$ times for each value of $B$, producing a distribution of absolute errors relative to the $B_{\max}$ estimate. Figure~\ref{fig:num_samples} summarizes how this error decreases as the number of subsamples grows, shown separately for synthetic and real-world datasets.

\vspace{-6mm}
\noindent Finally, to complement our complexity analysis (\S\ref{subsec:smm}), we examine how the runtime of $S_\mathrm{mM}$ scales with dataset size. We generate synthetic Gaussian data with fixed cluster structure ($k=5$, $d=10$), and increasing sample size $N$ (from $1,000$ to $500,000$), and compare a single-$k$ evaluation of $S_\mathrm{mM}$ (using $B=20$) with the standard (\texttt{scikit-learn}) Silhouette score computed on the full dataset. 

\begin{figure}[H]
    \centering
    \includegraphics[width=0.80\linewidth]{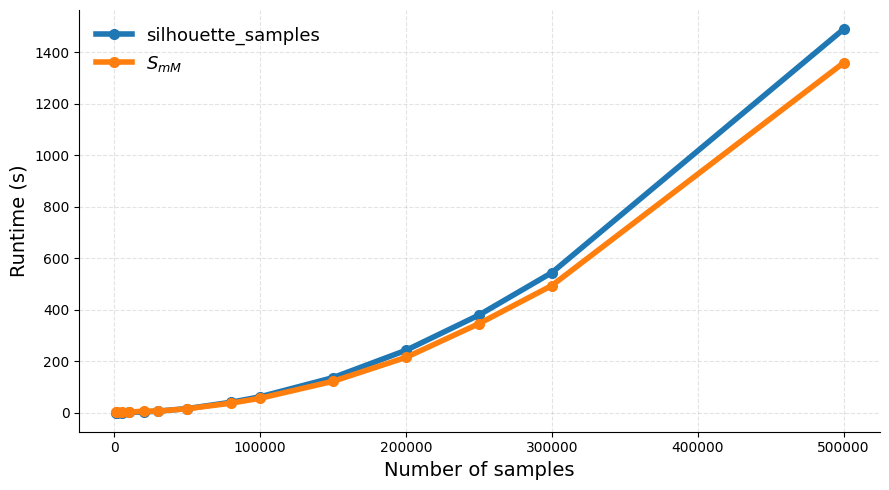}
    \caption{Runtime as a function of the number of samples $N$ for a single-$k$ evaluation of \textcolor{orange}{$S_\mathrm{mM}$} and for the \textcolor{MidnightBlue}{standard Silhouette score} computed on the full dataset.}
    \label{fig:runtime_scaling}
\end{figure}

\subsection{Results}\label{subsec:res}

Table~\ref{tab:main_kmeans_results} shows that $S_\mathrm{mM}$ (Eqs.~\ref{eq:smm},\ref{eq:max}) recovers the ground-truth number of clusters on all sixteen datasets considered (\S \ref{subsec:data}), making it the only criterion in our comparison to do so consistently. This holds across balanced synthetic data (\textsc{S1}, \textsc{S2}), strongly imbalanced and heterogeneous synthetic settings (\textsc{S3}, \textsc{S4}), tabular datasets, image embeddings, and text representations. Importantly, whenever one of the two subsample-averaged Silhouette views (Eq.~\ref{eq:subavg}) provides the stronger signal, $S_\mathrm{mM}$ follows it without requiring prior knowledge of whether micro- or macro-averaging should be preferred (Table~\ref{tab:main_kmeans_results}, Appendix~\ref{app:sec:val}); see, for example, \textsc{S1}, where both $\overline S_\mathrm{m}, \overline S_\mathrm{M}$ identify the ground-truth number of clusters, \textsc{S2}, where the micro view is more informative, and \textsc{S3} and \textsc{B77}, where the macro view is more informative (Table~\ref{tab:main_kmeans_results}, Appendix~\ref{app:sec:val}; Table~\ref{app:tab:supp_kgt_values}, Figs.~\ref{app:fig:trends1}--\ref{app:fig:trends6}). At the same time, $S_\mathrm{mM}$ remains effective in more ambiguous cases where neither view alone identifies $k_{\mathrm{GT}}$, such as \textsc{S4}, \textsc{Nsg}, and \textsc{Mds}. This behavior also extends beyond \emph{k}-means, as shown by the corresponding GMM, Bisecting \emph{k}-means in Appendix~\ref{app:sec:val} (Tables~\ref{app:tab:bisecting_main_results}--\ref{app:tab:gmm_main_results}). These findings directly support our motivating question: $S_\mathrm{mM}$ is able to combine the strengths of micro- and macro-averaged Silhouette for cluster-count selection without assuming in advance which aggregation strategy is preferable. This remains the case even in settings where neither view alone aligns with the ground truth, suggesting that the composite captures a more informative signal than either aggregation strategy on its own.

A second observation is that the subsample-averaged baselines $\overline S_\mathrm{m}$ and $\overline S_\mathrm{M}$ are generally more reliable than the corresponding averages across repeated full-data runs, which supports the use of repeated subsampled views rather than repeated random initializations alone. The automatic subsample-size rule (\S \ref{subsec:experiments}) is sufficient across all datasets and does not require dataset-specific tuning. Figure~\ref{fig:num_samples} further shows that the approximation error of $S_\mathrm{mM}$ decreases rapidly with the number of subsamples: the across-dataset median error is already below $0.01$ at $B=10$ for both synthetic and real data, while even the most difficult real datasets remain within roughly $0.05$ and improve steadily as $B$ increases. Thus, moderate values such as $B=20$--$30$ are adequate in practice. Table~\ref{app:tab:abl} in Appendix~\ref{app:sec:ablation} additionally shows that the $\tanh$ transformation yields the most reliable $S_\mathrm{mM}$ ground-truth recovery among the tested $z_b$ transformations (\S \ref{subsec:weighting}; Eq.~\ref{eq:zb}). Finally, Fig.~\ref{fig:runtime_scaling} shows that $S_\mathrm{mM}$ remains computationally competitive: despite including repeated subsampling, clustering, and aggregation, its runtime stays competitive with the full-data Silhouette baseline and the advantage becomes more pronounced as the dataset size grows.

\section{Conclusion}\label{sec:conc}

We introduced Composite Silhouette ($S_\mathrm{mM}$), a discrepancy-aware internal validation criterion that combines micro- and macro-averaged Silhouette scores through repeated subsampled clusterings. By using subsample-specific convex weights derived from the local disagreement between the two aggregation views, $S_\mathrm{mM}$ adaptively favors the more informative perspective without discarding the other. At the same time, the method remains computationally practical, with modest additional overhead beyond repeated subsampled Silhouette evaluation and natural compatibility with parallel computation.
Empirical results show that $S_\mathrm{mM}$ identifies the correct number of clusters more reliably than either subsample-averaged micro- or macro-averaged Silhouette alone, as well as standard internal baselines, across balanced, imbalanced, and structurally heterogeneous settings. These findings indicate that combining the two Silhouette views locally, rather than committing to one of them globally, yields a more robust basis for cluster-count selection.
A natural direction for future work is a more systematic study of the regimes in which the composite remains informative even when both component views are individually misleading, in order to better characterize the structural conditions that favor discrepancy-aware aggregation. Overall, $S_\mathrm{mM}$ provides a practical and broadly applicable framework for cluster-count selection that is adaptive and effective across diverse data domains.

\section*{Declarations}

\begin{itemize}
\item \textbf{Funding.} This work was supported by the Archimedes Research Unit, Athena Research Center, through the project ``ARCHIMEDES Unit: Research in Artificial Intelligence, Data Science, and Algorithms'', implemented within the framework of the National Recovery and Resilience Plan ``Greece 2.0'' and funded by the European Union -- NextGenerationEU.

\item \textbf{Disclosure of Interests.} The authors declare that they have no competing interests.
\end{itemize}

\bibliography{references}

\newpage

\begin{appendices}

\section{Extended Theoretical Analysis}\label{app:sec:theory}

\subsection{Property 1}\label{app:subsec:prop1}
\[
S_\mathrm{mM}^{(b)}
=
\frac{S_\mathrm{m}^{(b)}+S_\mathrm{M}^{(b)}}{2}
+
\frac{\Delta_b z_b}{2}
=
\frac{S_\mathrm{m}^{(b)}+S_\mathrm{M}^{(b)}}{2}
+
\frac{\Delta_b}{2}
\tanh\!\left(\frac{\Delta_b}{\Delta_{\max}+\varepsilon}\right).
\]

\medskip

\noindent \textbf{Proof.}
Starting from the definition of the subsample-specific composite score,
\[
S_\mathrm{mM}^{(b)}
= w_bS_\mathrm{m}^{(b)} + (1-w_b)S_\mathrm{M}^{(b)}.
\]
Rewriting,
\[
S_\mathrm{mM}^{(b)}
= w_b\bigl(S_\mathrm{m}^{(b)}-S_\mathrm{M}^{(b)}\bigr)+S_\mathrm{M}^{(b)}.
\]
Using \(\Delta_b=S_\mathrm{m}^{(b)}-S_\mathrm{M}^{(b)}\) and \(w_b=\frac{1+z_b}{2}\), we obtain
\[
S_\mathrm{mM}^{(b)}
= \frac{1+z_b}{2}\,\Delta_b + S_\mathrm{M}^{(b)}
= \frac{\Delta_b}{2} + \frac{\Delta_b z_b}{2} + S_\mathrm{M}^{(b)}.
\]
Since \(\Delta_b=S_\mathrm{m}^{(b)}-S_\mathrm{M}^{(b)}\),
\[
\frac{\Delta_b}{2}+S_\mathrm{M}^{(b)}
=
\frac{S_\mathrm{m}^{(b)}-S_\mathrm{M}^{(b)}}{2}+S_\mathrm{M}^{(b)}
=
\frac{S_\mathrm{m}^{(b)}+S_\mathrm{M}^{(b)}}{2}.
\]
Therefore,
\[
S_\mathrm{mM}^{(b)}
=
\frac{S_\mathrm{m}^{(b)}+S_\mathrm{M}^{(b)}}{2}
+
\frac{\Delta_b z_b}{2}.
\]
Finally, substituting
\[
z_b=\tanh\!\left(\frac{\Delta_b}{\Delta_{\max}+\varepsilon}\right)
\]
yields
\[
S_\mathrm{mM}^{(b)}
=
\frac{S_\mathrm{m}^{(b)}+S_\mathrm{M}^{(b)}}{2}
+
\frac{\Delta_b}{2}\tanh\!\left(\frac{\Delta_b}{\Delta_{\max}+\varepsilon}\right).
\]
\hfill $\square$

\subsection{Property 2}\label{app:subsec:prop2}

\[
S_\mathrm{mM}
=
\frac{\overline S_\mathrm{m}+\overline S_\mathrm{M}}{2}
+
\frac{1}{2B}\sum_{b=1}^B \Delta_b z_b.
\]

\medskip

\noindent \textbf{Proof.}
Starting from the definition of the Composite Silhouette score,
\[
S_\mathrm{mM}
=\frac{1}{B}\sum_{b=1}^{B}\left[w_bS_\mathrm{m}^{(b)}+(1-w_b)S_\mathrm{M}^{(b)}\right].
\]
Using the subsample-level identity
\[
w_bS_\mathrm{m}^{(b)}+(1-w_b)S_\mathrm{M}^{(b)}
=
\frac{S_\mathrm{m}^{(b)}+S_\mathrm{M}^{(b)}}{2}
+
\frac{\Delta_b z_b}{2},
\]
we obtain
\[
S_\mathrm{mM}
=
\frac{1}{B}\sum_{b=1}^{B}
\left[
\frac{S_\mathrm{m}^{(b)}+S_\mathrm{M}^{(b)}}{2}
+
\frac{\Delta_b z_b}{2}
\right].
\]
Distributing the average gives
\[
S_\mathrm{mM}
=
\frac{1}{2B}\sum_{b=1}^{B}\left(S_\mathrm{m}^{(b)}+S_\mathrm{M}^{(b)}\right)
+
\frac{1}{2B}\sum_{b=1}^{B}\Delta_b z_b.
\]
By the definitions
\[
\overline S_\mathrm{m}=\frac{1}{B}\sum_{b=1}^{B}S_\mathrm{m}^{(b)},
\qquad
\overline S_\mathrm{M}=\frac{1}{B}\sum_{b=1}^{B}S_\mathrm{M}^{(b)},
\]
the first term becomes
\[
\frac{1}{2B}\sum_{b=1}^{B}\left(S_\mathrm{m}^{(b)}+S_\mathrm{M}^{(b)}\right)
=
\frac{\overline S_\mathrm{m}+\overline S_\mathrm{M}}{2}.
\]
Therefore,
\[
S_\mathrm{mM}
=
\frac{\overline S_\mathrm{m}+\overline S_\mathrm{M}}{2}
+
\frac{1}{2B}\sum_{b=1}^B \Delta_b z_b.
\]
\hfill $\square$

\medskip

\noindent \textbf{Probabilistic setup.}
For the probabilistic results below, we consider a fixed candidate number of clusters $k\in\mathcal K$. The subsamples are assumed to be drawn independently according to the sampling scheme, while any internal randomness of the clustering algorithm is also independent across trials and independent of the subsampling. Since, in our implemented method, the normalizing quantity $\Delta_{\max}(k)$ is computed from the same subsamples used to form $S_\mathrm{mM}(k)$, the resulting terms are not independent. To obtain concentration bounds with standard tools, we therefore analyze a closely related version in which $\Delta_{\max}(k)$ is estimated from an independent set of subsamples and then used to form the final score. This preserves the form of the method while ensuring that the per-subsample composite scores used in the analysis are independent and identically distributed. In practice, this modification is mainly technical. The quantity $\Delta_{\max}(k)$ enters the method only through the normalization of the discrepancies, so it acts as a relative scale parameter rather than as a source of structural information on its own. Consequently, replacing the empirical $\Delta_{\max}(k)$ computed from the same subsamples by an independent estimate does not alter the form of the weighting rule, but only its normalization. When $B$ is moderate to large, both quantities are expected to provide similar scaling of the discrepancies, and therefore to induce similar normalized values, weights, and final composite scores. The two versions are thus introduced to separate dependence for the sake of analysis, rather than because they represent different procedures.

\subsection{Fixed-$k$ concentration}\label{app:subsec:fixedk}

Fix $k\in\mathcal K$, and define
\[
\mu(k)=\mathbb E\!\left[S_\mathrm{mM}^{(b)}(k)\mid \Delta_{\max}(k)\right].
\]
Then, for any $t>0$,
\[
\mathbb P\!\left(\left|S_\mathrm{mM}(k)-\mu(k)\right|\ge t \,\middle|\, \Delta_{\max}(k)\right)
\le
2\exp\!\left(-\frac{Bt^2}{2}\right).
\]
Equivalently, for any $\delta\in(0,1)$, with conditional probability at least $1-\delta$,
\[
\left|S_\mathrm{mM}(k)-\mu(k)\right|
\le
\sqrt{\frac{2\ln(2/\delta)}{B}}.
\]

\medskip

\noindent \textbf{Proof.}
Under our setup, the random variables
\[
S_\mathrm{mM}^{(1)}(k),\dots,S_\mathrm{mM}^{(B)}(k)
\]
are independent and identically distributed conditional on $\Delta_{\max}(k)$. Each of them lies in $[-1,1]$. Hoeffding's inequality for bounded independent random variables therefore gives
\[
\mathbb P\!\left(\left|\frac{1}{B}\sum_{b=1}^{B}S_\mathrm{mM}^{(b)}(k)-\mu(k)\right|\ge t \,\middle|\, \Delta_{\max}(k)\right)
\le
2\exp\!\left(-\frac{Bt^2}{2}\right),
\]
which is exactly the desired bound since
\[
S_\mathrm{mM}(k)=\frac{1}{B}\sum_{b=1}^{B}S_\mathrm{mM}^{(b)}(k).
\]
Solving
\[
2\exp\!\left(-\frac{Bt^2}{2}\right)=\delta
\]
for $t$ yields
\[
t=\sqrt{\frac{2\ln(2/\delta)}{B}},
\]
which proves the claim.
\hfill $\square$

\subsection{Uniform concentration over the candidate set}\label{app:subsec:uniform}

Assume that the candidate set $\mathcal K$ is finite. Then, for any $\delta\in(0,1)$,
\[
\sup_{k\in\mathcal K}\left|S_\mathrm{mM}(k)-\mu(k)\right|
\le
\sqrt{\frac{2\ln(2|\mathcal K|/\delta)}{B}}
\]
with conditional probability at least $1-\delta$.

\medskip

\noindent \textbf{Proof.}
Fix $t>0$. By the fixed-$k$ concentration result above, for each $k\in\mathcal K$,
\[
\mathbb P\!\left(\left|S_\mathrm{mM}(k)-\mu(k)\right|\ge t \,\middle|\, \Delta_{\max}(k)\right)
\le
2\exp\!\left(-\frac{Bt^2}{2}\right).
\]
Applying the union bound over $k\in\mathcal K$ gives
\[
\mathbb P\!\left(\exists\,k\in\mathcal K:\left|S_\mathrm{mM}(k)-\mu(k)\right|\ge t \,\middle|\, \Delta_{\max}(k)\right)
\le
2|\mathcal K|\exp\!\left(-\frac{Bt^2}{2}\right).
\]
Setting the right-hand side equal to $\delta$ and solving for $t$ yields
\[
t=\sqrt{\frac{2\ln(2|\mathcal K|/\delta)}{B}},
\]
which proves the claim.
\hfill $\square$

\subsection{Recovery guarantee under a margin condition}\label{app:subsec:recovery}

Let
\[
k^\dagger\in\arg\max_{k\in\mathcal K}\mu(k)
\]
be an optimal candidate under the subsampling objective, and assume that this maximizer is unique with positive margin
\[
\gamma:=\mu(k^\dagger)-\max_{k\in\mathcal K,\;k\neq k^\dagger}\mu(k)>0.
\]
Let
\[
k^\star\in\arg\max_{k\in\mathcal K}S_\mathrm{mM}(k)
\]
denote the maximizer of the empirical Composite Silhouette score. If
\[
B\ge \frac{8}{\gamma^2}\ln\!\left(\frac{2|\mathcal K|}{\delta}\right),
\]
then
\[
\mathbb P\!\left(k^\star=k^\dagger \,\middle|\, \Delta_{\max}(k)\right)\ge 1-\delta.
\]

\medskip

\noindent \textbf{Proof.}
By A.2, with conditional probability at least $1-\delta$,
\[
\sup_{k\in\mathcal K}\left|S_\mathrm{mM}(k)-\mu(k)\right|
\le
\sqrt{\frac{2\ln(2|\mathcal K|/\delta)}{B}}.
\]
Suppose this event holds and let
\[
\eta:=\sqrt{\frac{2\ln(2|\mathcal K|/\delta)}{B}}.
\]
Then
\[
S_\mathrm{mM}(k^\dagger)\ge \mu(k^\dagger)-\eta,
\]
while for every $k\neq k^\dagger$,
\[
S_\mathrm{mM}(k)\le \mu(k)+\eta\le \mu(k^\dagger)-\gamma+\eta.
\]
Hence, if $\eta\le \gamma/2$, then
\[
S_\mathrm{mM}(k^\dagger)\ge \mu(k^\dagger)-\eta\ge \mu(k^\dagger)-\gamma+\eta\ge S_\mathrm{mM}(k)
\]
for all $k\neq k^\dagger$, so $k^\dagger$ is the unique maximizer of $S_\mathrm{mM}(k)$. The condition $\eta\le \gamma/2$ is equivalent to
\[
\sqrt{\frac{2\ln(2|\mathcal K|/\delta)}{B}}\le \frac{\gamma}{2},
\]
which in turn gives
\[
B\ge \frac{8}{\gamma^2}\ln\!\left(\frac{2|\mathcal K|}{\delta}\right).
\]
Therefore, under this condition,
\[
\mathbb P\!\left(k^\star=k^\dagger \,\middle|\, \Delta_{\max}(k)\right)\ge 1-\delta.
\]
\hfill $\square$

\section{Transformations Ablation}\label{app:sec:ablation} 

We examine the sensitivity of Composite Silhouette to the choice of transformation applied to the normalized micro--macro discrepancy $\widetilde{\Delta}_b$ before constructing the subsample-specific weight, while keeping the remaining pipeline fixed (including subsampling, $k$-means clustering, and Silhouette computation). In the main method, we use
$z_b=\tanh(\widetilde{\Delta}_b)$ (Eq.~\ref{eq:zb}),
which yields the weight $w_b=(1+z_b)/2$ (Eq.~\ref{eq:wb}). To assess whether this choice is important in practice, we compare the proposed $\tanh$ transformation against three alternatives: a linear mapping, $z_b=\widetilde{\Delta}_b$; a sigmoid transformation, implemented as $w_b=(1+\exp(-\alpha \widetilde{\Delta}_b))^{-1}$ with $\alpha=1$; and a step-based transformation, which assigns full weight to the micro-averaged Silhouette when $\widetilde{\Delta}_b>0$ and full weight to the macro-averaged Silhouette otherwise. For each variant, we recompute $S_{\mathrm{mM}}$ and report the resulting selected number of clusters across the benchmark datasets.

\noindent Table~\ref{app:tab:abl} shows that the proposed $\tanh$ transformation is the most stable and accurate choice overall. In particular, it recovers the correct number of clusters on all datasets reported, whereas the alternative transformations yield several under- and over-estimations. This empirical pattern supports the role of $\tanh$ as a principled compromise: unlike the linear mapping, it smoothly compresses large discrepancies and thus avoids excessive sensitivity to extreme values; unlike the step rule, it preserves gradual adjustments in the relative weighting between micro- and macro-averaged Silhouette scores. We note that a sigmoid transformation with $\alpha=2$ would be mathematically equivalent to the $\tanh$ weighting used in the main method, so in this ablation we use $\alpha=1$ to obtain a genuinely distinct smooth alternative.

\begin{table}[h]
\centering
\caption{Estimated numbers of clusters via $S_\mathrm{mM}$ using various $z_b$ transformations (original tanh, linear, sigmoid, step). \textbf{Bold values} indicate correct number of clusters selection, \textbf{\textcolor{red}{red values}} indicate suboptimal selection.}\label{app:tab:abl}
\begin{tabular}{lrrrr}
\toprule
& \multicolumn{4}{c}{$S_{\mathrm{mM}}$ $k$ selection with $z_b$ transformation} \\
\cmidrule(lr){2-5}
\textbf{Dataset } & \textbf{ tanh }  &  \textbf{ linear }  &  \textbf{ sigmoid }  &  \textbf{ step (sign) }  \\
\midrule
\textsc{S1}  & \textbf{5}  & \textbf{5}  & \textbf{5}  & \textbf{5}  \\
\textsc{S2}  & \textbf{6}  & \textbf{\textcolor{red}{3}}  & \textbf{6}  & \textbf{\textcolor{red}{3}}  \\
\textsc{S3}  & \textbf{5}  & \textbf{5}  & \textbf{5}  & \textbf{5}  \\
\textsc{S4}  & \textbf{12} & \textbf{\textcolor{red}{13}} & \textbf{12} & \textbf{\textcolor{red}{13}} \\
\midrule
\textsc{Pks} & \textbf{2}  & \textbf{2}  & \textbf{\textcolor{red}{3}}  & \textbf{2}  \\
\textsc{Wne} & \textbf{3}  & \textbf{3}  & \textbf{\textcolor{red}{4}}  & \textbf{3}  \\
\textsc{Htr} & \textbf{2}  & \textbf{2}  & \textbf{2}  & \textbf{2}  \\
\textsc{Dgt} & \textbf{10} & \textbf{\textcolor{red}{8}} & \textbf{10} & \textbf{10} \\
\textsc{Bnk} & \textbf{2}  & \textbf{2}  & \textbf{2}  & \textbf{2}  \\
\textsc{Nsg} & \textbf{20} & \textbf{20} & \textbf{20} & \textbf{\textcolor{red}{21}} \\
\textsc{Spm} & \textbf{2}  & \textbf{2}  & \textbf{2}  & \textbf{2}  \\
\textsc{Stl} & \textbf{10} & \textbf{10} & \textbf{10} & \textbf{\textcolor{red}{8}} \\
\textsc{Bbc} & \textbf{5}  & \textbf{5}  & \textbf{5}  & \textbf{\textcolor{red}{6}}  \\
\textsc{Bld} & \textbf{2}  & \textbf{\textcolor{red}{3}}  & \textbf{2}  & \textbf{2}  \\
\textsc{Mds} & \textbf{14} & \textbf{14} & \textbf{14} & \textbf{\textcolor{red}{15}} \\
\textsc{B77} & \textbf{77} & \textbf{\textcolor{red}{72}} & \textbf{77} & \textbf{77} \\
\bottomrule
\end{tabular}
\end{table}

\newpage

\section{Extended Empirical Validation}\label{app:sec:val}

\begin{table*}[hbpt]
\centering
\setlength{\tabcolsep}{4pt}
\caption{Values attained at the ground-truth number of clusters $k_{\mathrm{GT}}$ under the $k$-means protocol. Higher values indicate better performance for $S_\mathrm{mM}$, $\overline S_\mathrm{m}$, $\overline S_\mathrm{M}$, avg $S_\mathrm{m}$, avg $S_\mathrm{M}$, and avg CH, whereas lower values indicate better performance for avg DB.}
\label{app:tab:supp_kgt_values}
\begin{tabular}{lccccccc}
\toprule
Dataset & $S_\mathrm{mM}$ & $\overline S_\mathrm{m}$ & $\overline S_\mathrm{M}$ & avg $S_\mathrm{m}$ & avg $S_\mathrm{M}$ & avg CH & avg DB \\
\midrule
\textsc{S1}  & \textbf{0.8728} & \textbf{0.8727} & \textbf{0.8728} & \textbf{0.8728} & \textbf{0.8728} & \textbf{284172.5} & \textbf{0.1781} \\
\textsc{S2}  & \textbf{0.6924} & \textbf{0.6924} & 0.6923 & \textbf{0.6924} & 0.6924 & \textbf{62284.8}  &\textbf{0.4289} \\
\textsc{S3}  & \textbf{0.6703} & 0.4767 & \textbf{0.7013} & 0.4851 & \textbf{0.7200} & \textbf{3879.3}   & \textbf{0.4852} \\
\textsc{S4}  & \textbf{0.6114} & 0.4236 & 0.6448 & 0.4211 & 0.6540 & 7214.3   & 0.5013 \\
\midrule
\textsc{Pks} & \textbf{0.4047} & \textbf{0.4195} & \textbf{0.3221} & \textbf{0.4273} & \textbf{0.3287} & 99.7     & \textbf{1.0917} \\
\textsc{Wne} & \textbf{0.2875} & \textbf{0.2835} & \textbf{0.2887} & \textbf{0.2835} & \textbf{0.2886} & \textbf{70.6}     & \textbf{1.3918} \\
\textsc{Bld} & \textbf{0.4186} & \textbf{0.4277} & 0.3681 & \textbf{0.4315} & 0.3689 & 429.4    & 1.0225 \\
\textsc{Dgt} & \textbf{0.1848} & \textbf{0.1800} & \textbf{0.1861} & \textbf{0.1776} & 0.1819 & 165.0    & 1.9596 \\
\textsc{Bbc} & \textbf{0.1111} & 0.1084 & \textbf{0.1114} & \textbf{0.1081} & \textbf{0.1109} & 160.9    & 2.6293 \\
\textsc{Htr} & \textbf{0.5874} & \textbf{0.6128} & \textbf{0.4444} & \textbf{0.6147} & \textbf{0.4509} & 9821.5   & 0.9174 \\
\textsc{Stl} & \textbf{0.0292} & \textbf{0.0303} & \textbf{0.0244} & 0.0303 & 0.0225 & 136.3    & \textbf{3.6423} \\
\textsc{Nsg} & \textbf{0.1996} & 0.0853 & 0.2222 & 0.0778 & 0.2019 & 321.0    & 1.8722 \\
\textsc{Spm}  & \textbf{0.4320} & \textbf{0.4508} & \textbf{0.3658} & \textbf{0.3760} & \textbf{0.3084} & 245.1   & \textbf{2.0751} \\
\textsc{Mds} & \textbf{0.2689} & 0.2680 & 0.2681 & 0.2657 & \textbf{0.2654} & \textbf{45.6}     & \textbf{1.7025} \\
\textsc{Bnk} & \textbf{0.2325} & \textbf{0.2409} & \textbf{0.2091} & \textbf{0.2980} & \textbf{0.2314} & 6315.7   & 1.7660 \\
\textsc{B77} & \textbf{0.1720} & 0.1674 & \textbf{0.1736} & 0.1679 & \textbf{0.1758} & 281.4    & 1.9576 \\
\bottomrule
\end{tabular}
\end{table*}

\begin{table*}[h]
\centering
\setlength{\tabcolsep}{4pt}
\caption{Selected number of clusters obtained via $S_\mathrm{mM}$, $\overline S_\mathrm{m}$, $\overline S_\mathrm{M}$, avg $S_\mathrm{m}$, avg $S_\mathrm{M}$, avg CH, and avg DB under the Bisecting $k$-means protocol. \textbf{Bold} entries indicate that the selected $k$ matches the ground-truth $k_\mathrm{GT}$.}
\vspace{0.5mm}
\label{app:tab:bisecting_main_results}
\begin{tabular}{lccccccc}
\toprule
Dataset & $S_\mathrm{mM}$ & $\overline S_\mathrm{m}$ & $\overline S_\mathrm{M}$ & avg $S_\mathrm{m}$ & avg $S_\mathrm{M}$ & avg CH & avg DB \\
\midrule
\textsc{S1}  & \textbf{5}  & \textbf{5}  & \textbf{5}  & \textbf{5}  & \textbf{5}  & \textbf{5}  & \textbf{5} \\
\textsc{S2}  & \textbf{6}  & \textbf{6}  & 3           & \textbf{6}  & 3           & \textbf{6}  & \textbf{6} \\
\textsc{S3}  & 3           & 2           & \textbf{5}  & 3           & 3           & 10          & \textbf{5} \\
\textsc{S4}  & 11          & 8           & 8           & 8           & 8           & 16          & 8 \\
\midrule
\textsc{Pks} & \textbf{2}  & \textbf{2}  & \textbf{2}  & \textbf{2}  & \textbf{2}  & \textbf{2}  & \textbf{2} \\
\textsc{Wne} & \textbf{3}  & \textbf{3}  & \textbf{3}  & 2           & 2           & 2           & \textbf{3} \\
\textsc{Bld} & \textbf{2}  & \textbf{2}  & 4           & \textbf{2}  & 4           & 4           & 4 \\
\textsc{Dgt} & \textbf{10} & \textbf{10} & \textbf{10} & \textbf{10} & 9           & 2           & \textbf{10} \\
\textsc{Bbc} & \textbf{5}  & \textbf{5}  & \textbf{5}  & \textbf{5}  & \textbf{5}  & 2           & \textbf{5} \\
\textsc{Htr} & \textbf{2}  & \textbf{2}  & \textbf{2}  & \textbf{2}  & \textbf{2}  & \textbf{2}  & 4 \\
\textsc{Stl} & \textbf{10} & \textbf{10} & 9           & 12          & 8           & 6           & 8 \\
\textsc{Nsg} & 19          & 15          & 19          & 15          & 23          & 15          & 24 \\
\textsc{Spm}  & \textbf{2}  & \textbf{2}  & \textbf{2}  & \textbf{2}  & \textbf{2}  & 3           & 7 \\
\textsc{Mds} & 15          & 15          & \textbf{14} & 16          & 13          & 13          & \textbf{14} \\
\textsc{Bnk} & \textbf{2}  & \textbf{2}  & \textbf{2}  & \textbf{2}  & \textbf{2}  & 3           & 8 \\
\textsc{B77} & 72          & 72          & 72          & 72          & 72          & 72          & 82 \\
\bottomrule
\end{tabular}
\end{table*}

\begin{table*}[t]
\centering
\setlength{\tabcolsep}{4pt}
\caption{Selected number of clusters obtained via $S_\mathrm{mM}$, $\overline S_\mathrm{m}$, $\overline S_\mathrm{M}$, avg $S_\mathrm{m}$, avg $S_\mathrm{M}$, avg CH, and avg DB under the Gaussian mixture model protocol. \textbf{Bold} entries indicate that the selected $k$ matches the ground-truth $k_\mathrm{GT}$.}
\vspace{0.5mm}
\label{app:tab:gmm_main_results}
\begin{tabular}{lccccccc}
\toprule
Dataset & $S_\mathrm{mM}$ & $\overline S_\mathrm{m}$ & $\overline S_\mathrm{M}$ & avg $S_\mathrm{m}$ & avg $S_\mathrm{M}$ & avg CH & avg DB \\
\midrule
\textsc{S1}  & \textbf{5}  & \textbf{5}  & \textbf{5}  & \textbf{5}  & \textbf{5}  & \textbf{5}  & \textbf{5} \\
\textsc{S2}  & \textbf{6}  & \textbf{6}  & 3           & \textbf{6}  & 3           & \textbf{6}  & \textbf{6} \\
\textsc{S3}  & 3           & 3           & 3           & 3           & \textbf{5}  & 10          & \textbf{5} \\
\textsc{S4}  & 11          & 8           & 9           & 9           & 9           & 16          & 9 \\
\midrule
\textsc{Pks} & \textbf{2}  & \textbf{2}  & \textbf{2}  & \textbf{2}  & \textbf{2}  & 3           & \textbf{2} \\
\textsc{Wne} & \textbf{3}  & \textbf{3}  & \textbf{3}  & \textbf{3}  & \textbf{3}  & \textbf{3}  & \textbf{3} \\
\textsc{Bld} & \textbf{2}  & \textbf{2}  & \textbf{2}  & 5           & \textbf{2}  & 5           & \textbf{2} \\
\textsc{Dgt} & \textbf{10} & \textbf{10} & \textbf{10} & \textbf{10} & 9           & 6           & \textbf{10} \\
\textsc{Bbc} & \textbf{5}  & 6           & \textbf{5}  & \textbf{5}  & \textbf{5}  & 2           & 7 \\
\textsc{Htr} & \textbf{2}  & \textbf{2}  & \textbf{2}  & \textbf{2}  & \textbf{2}  & 3           & 3 \\
\textsc{Stl} & \textbf{10} & \textbf{10} & \textbf{10} & 6           & 6           & 6           & \textbf{10} \\
\textsc{Nsg} & 24          & 24          & 24          & 24          & 24          & 24          & 24 \\
\textsc{Spm}  & \textbf{2}  & \textbf{2}  & \textbf{2}  & \textbf{2}  & \textbf{2}  & \textbf{2}  & \textbf{2} \\
\textsc{Mds} & \textbf{14} & 15          & 12          & 15          & \textbf{14} & \textbf{14} & \textbf{14} \\
\textsc{Bnk} & \textbf{2}  & \textbf{2}  & \textbf{2}  & \textbf{2}  & \textbf{2}  & \textbf{2}  & \textbf{2} \\
\textsc{B77} & \textbf{77} & 82          & \textbf{77} & 81          & \textbf{77} & 72          & 81 \\
\bottomrule
\end{tabular}
\end{table*}

\begin{figure}[t]
    \centering

    \begin{subfigure}[t]{0.95\textwidth}
        \centering
        \includegraphics[width=\textwidth]{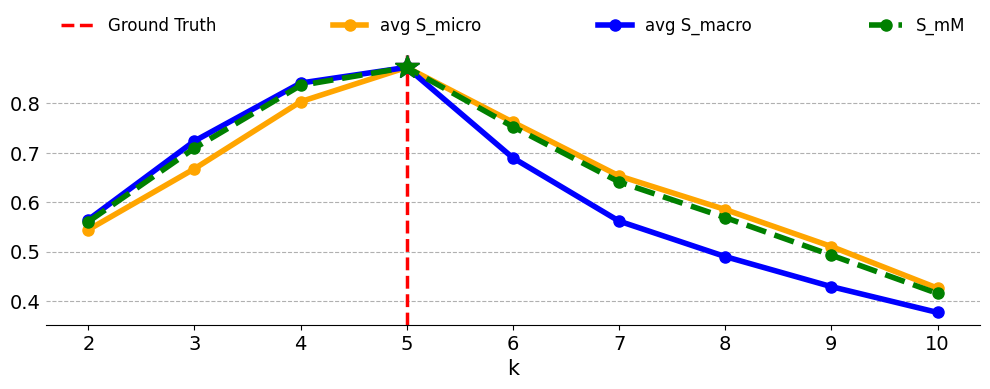}
        \caption{S1}
        \label{fig:s1trends}
    \end{subfigure}

    \vspace{0.8em}

    \begin{subfigure}[t]{0.95\textwidth}
        \centering
        \includegraphics[width=\textwidth]{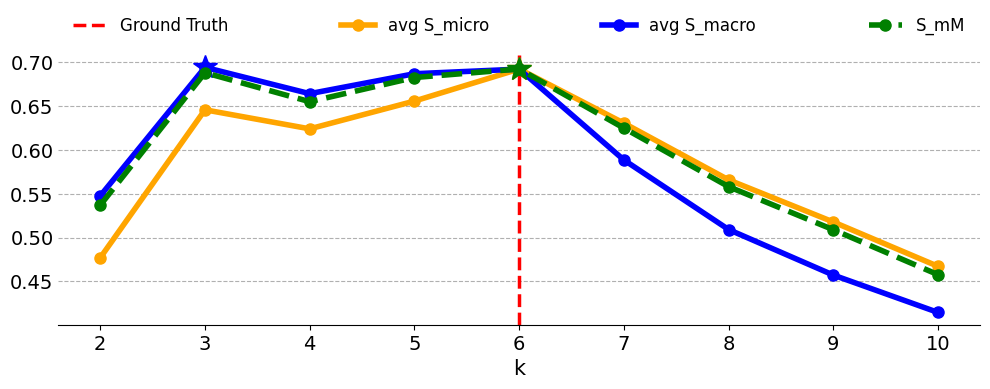}
        \caption{S2}
        \label{fig:s2trends}
    \end{subfigure}

    \vspace{0.8em}

    \begin{subfigure}[t]{0.95\textwidth}
        \centering
        \includegraphics[width=\textwidth]{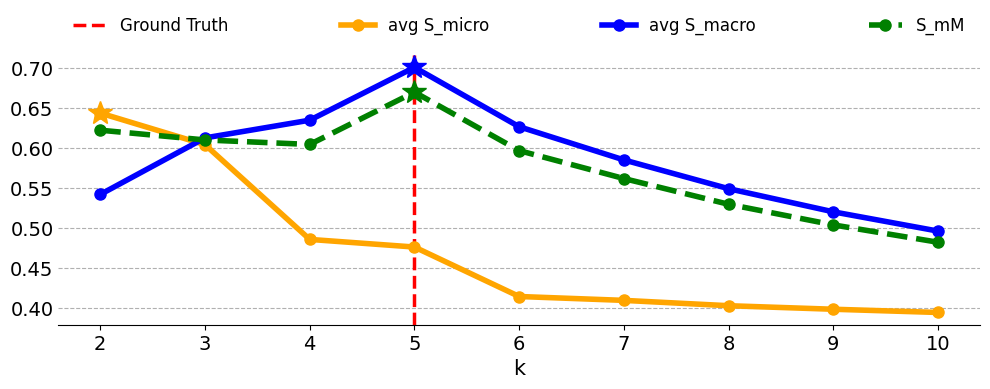}
        \caption{S3}
        \label{fig:s3trends}
    \end{subfigure}

    \caption{Trends of \textcolor{OliveGreen}{$S_\mathrm{mM}$}, \textcolor{orange}{$\overline S_\mathrm{m}$}, and \textcolor{blue}{$\overline S_\mathrm{M}$} over the candidate numbers of clusters $k$. Stars ($\star$) denote the optimal cluster count selected by each criterion. The \textcolor{red}{ dashed line} denotes the ground-truth number of clusters. Datasets: S1, S2, and S3.}
    \label{app:fig:trends1}
\end{figure}

\begin{figure}[t]
    \centering

    \begin{subfigure}[t]{0.95\textwidth}
        \centering
        \includegraphics[width=\textwidth]{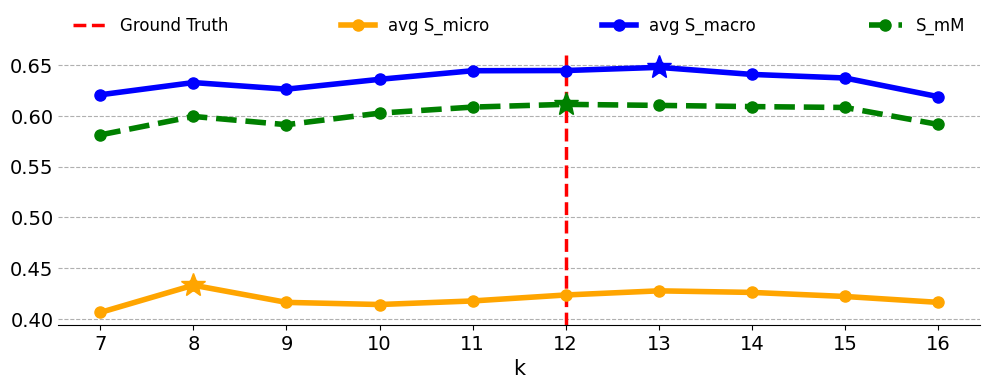}
        \caption{S4}
        \label{fig:s4trends}
    \end{subfigure}

    \vspace{0.8em}

    \begin{subfigure}[t]{0.95\textwidth}
        \centering
        \includegraphics[width=\textwidth]{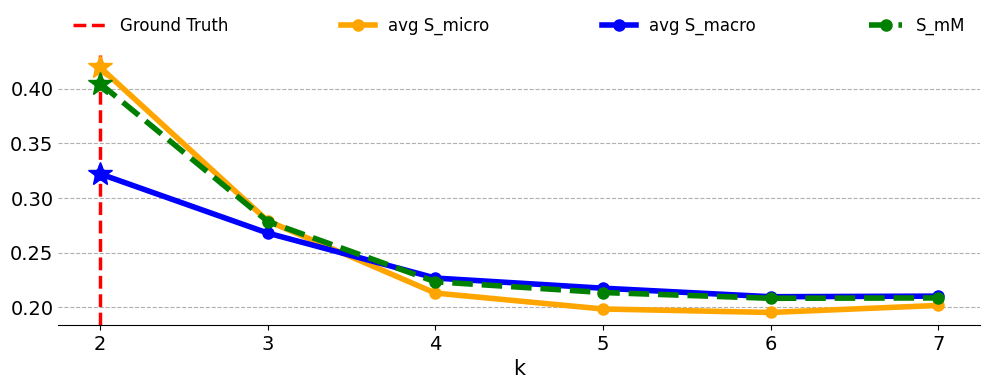}
        \caption{\textsc{Pks}}
        \label{fig:pkstrends}
    \end{subfigure}

    \vspace{0.8em}

    \begin{subfigure}[t]{0.95\textwidth}
        \centering
        \includegraphics[width=\textwidth]{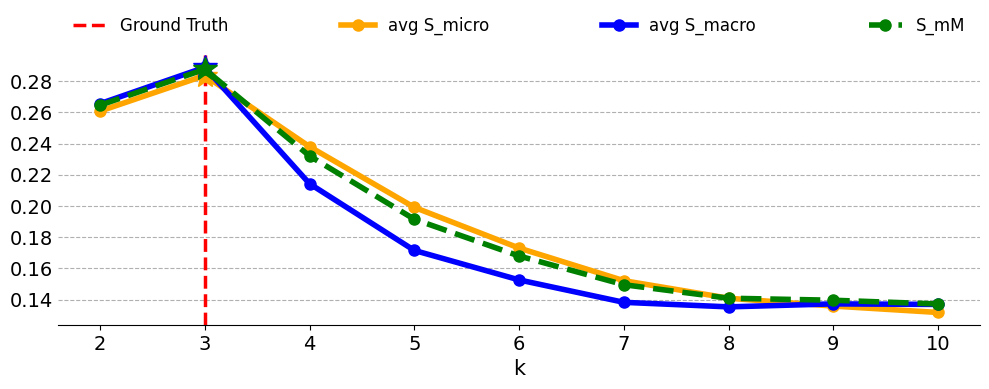}
        \caption{\textsc{Wne}}
        \label{fig:wnetrends}
    \end{subfigure}

    \caption{Trends of \textcolor{OliveGreen}{$S_\mathrm{mM}$}, \textcolor{orange}{$\overline S_\mathrm{m}$}, and \textcolor{blue}{$\overline S_\mathrm{M}$} over the candidate numbers of clusters $k$. Stars ($\star$) denote the optimal cluster count selected by each criterion. The \textcolor{red}{ dashed line} denotes the ground-truth number of clusters. Datasets: S4, \textsc{Pks}, and \textsc{Wne}.}
    \label{app:fig:trends2}
\end{figure}

\begin{figure}[t]
    \centering

    \begin{subfigure}[t]{0.95\textwidth}
        \centering
        \includegraphics[width=\textwidth]{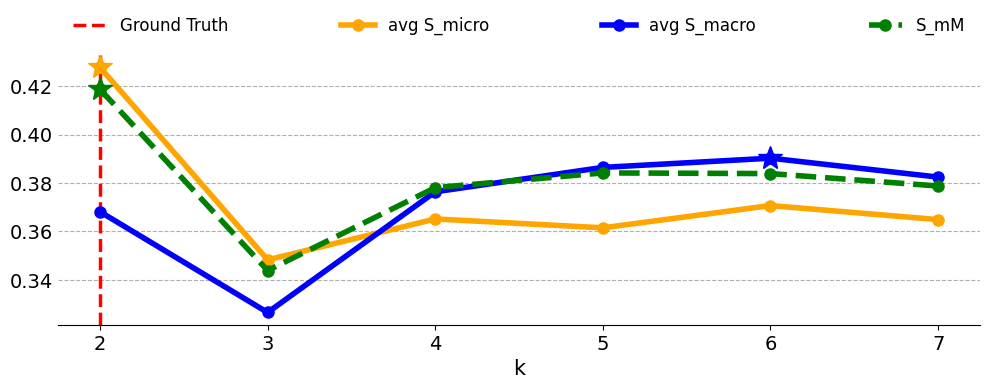}
        \caption{\textsc{Bld}}
        \label{fig:bldtrends}
    \end{subfigure}

    \vspace{0.8em}

    \begin{subfigure}[t]{0.95\textwidth}
        \centering
        \includegraphics[width=\textwidth]{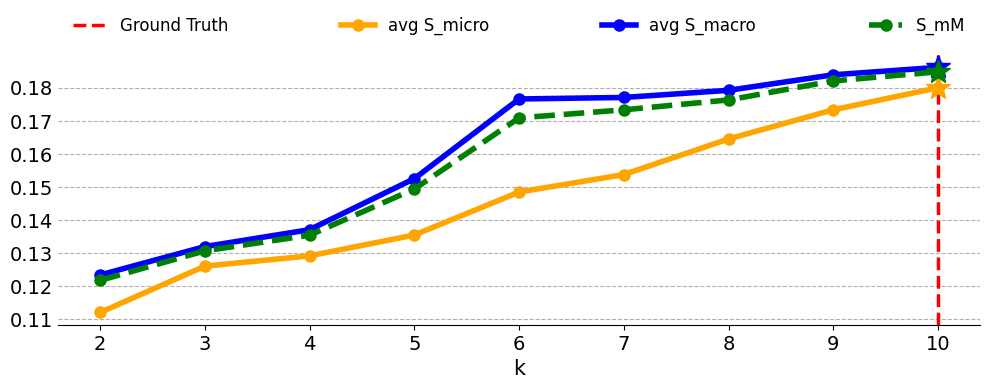}
        \caption{\textsc{Dig}}
        \label{fig:digtrends}
    \end{subfigure}

    \vspace{0.8em}

    \begin{subfigure}[t]{0.95\textwidth}
        \centering
        \includegraphics[width=\textwidth]{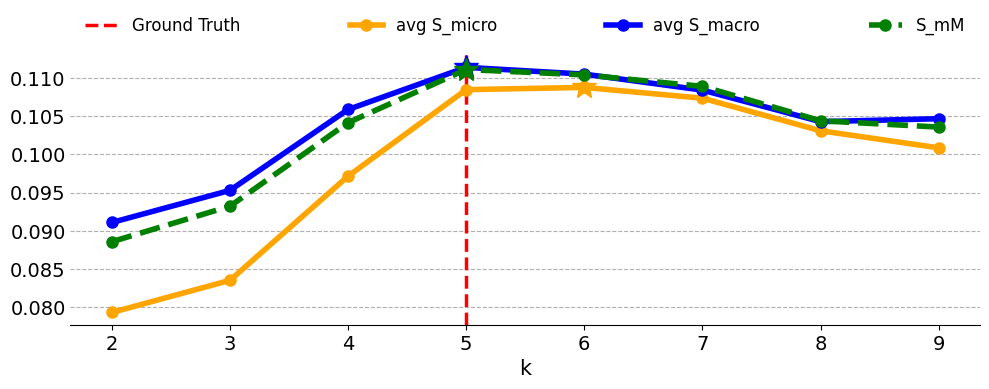}
        \caption{\textsc{Bbc}}
        \label{fig:bbctrends}
    \end{subfigure}

    \caption{Trends of \textcolor{OliveGreen}{$S_\mathrm{mM}$}, \textcolor{orange}{$\overline S_\mathrm{m}$}, and \textcolor{blue}{$\overline S_\mathrm{M}$} over the candidate numbers of clusters $k$. Stars ($\star$) denote the optimal cluster count selected by each criterion. The \textcolor{red}{ dashed line} denotes the ground-truth number of clusters. Datasets: \textsc{Bld}, \textsc{Dig}, and \textsc{Bbc}.}
    \label{app:fig:trends3}
\end{figure}

\begin{figure}[t]
    \centering

    \begin{subfigure}[t]{0.95\textwidth}
        \centering
        \includegraphics[width=\textwidth]{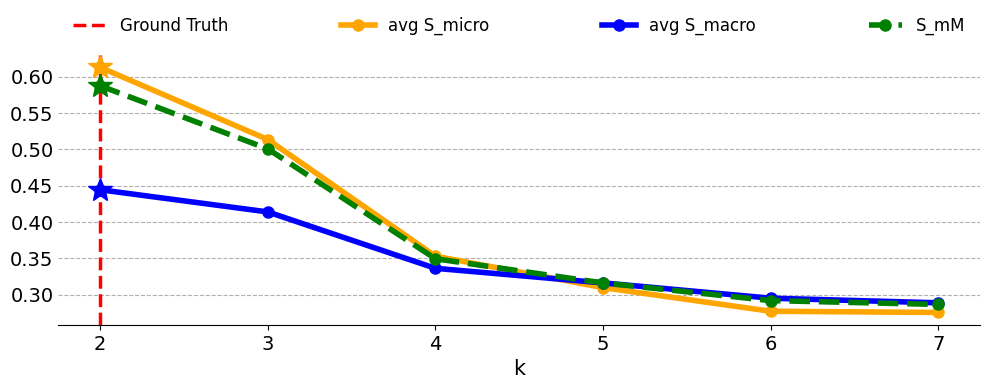}
        \caption{\textsc{Htr}}
        \label{fig:htrtrends}
    \end{subfigure}

    \vspace{0.8em}

    \begin{subfigure}[t]{0.95\textwidth}
        \centering
        \includegraphics[width=\textwidth]{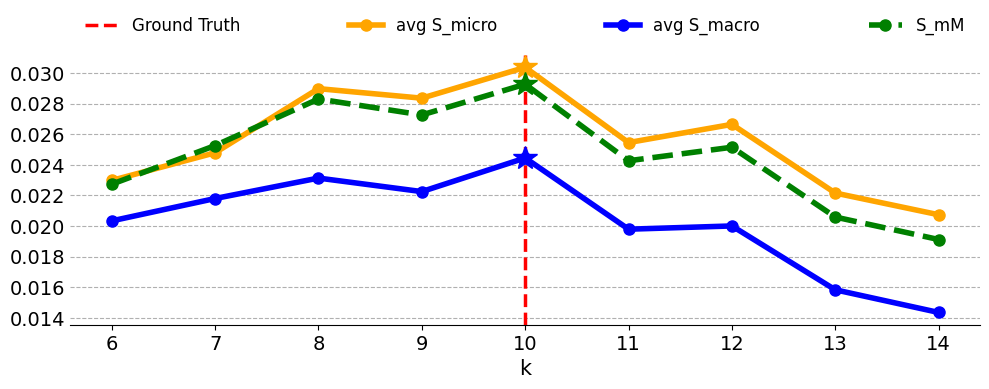}
        \caption{\textsc{Stl}}
        \label{fig:stltrends}
    \end{subfigure}

    \vspace{0.8em}

    \begin{subfigure}[t]{0.95\textwidth}
        \centering
        \includegraphics[width=\textwidth]{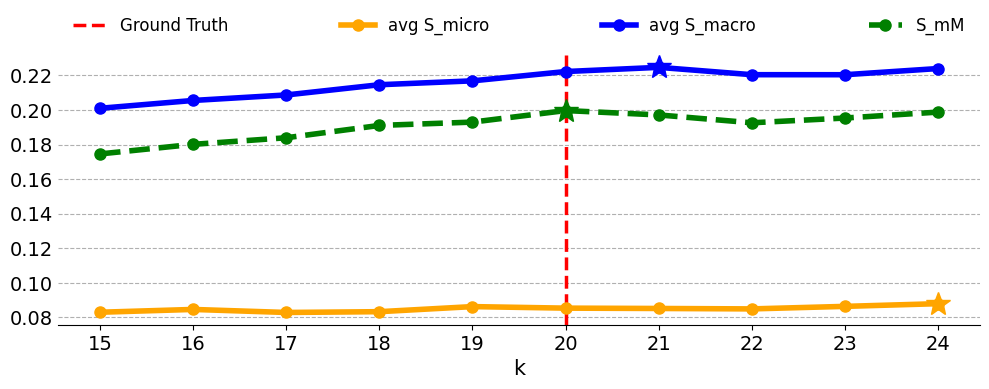}
        \caption{\textsc{Nsg}}
        \label{fig:nsgtrends}
    \end{subfigure}

    \caption{Trends of \textcolor{OliveGreen}{$S_\mathrm{mM}$}, \textcolor{orange}{$\overline S_\mathrm{m}$}, and \textcolor{blue}{$\overline S_\mathrm{M}$} over the candidate numbers of clusters $k$. Stars ($\star$) denote the optimal cluster count selected by each criterion. The \textcolor{red}{ dashed line} denotes the ground-truth number of clusters. Datasets: \textsc{Htr}, \textsc{Stl}, and \textsc{Nsg}.}
    \label{app:fig:trends4}
\end{figure}

\begin{figure}[t]
    \centering

    \begin{subfigure}[t]{0.95\textwidth}
        \centering
        \includegraphics[width=\textwidth]{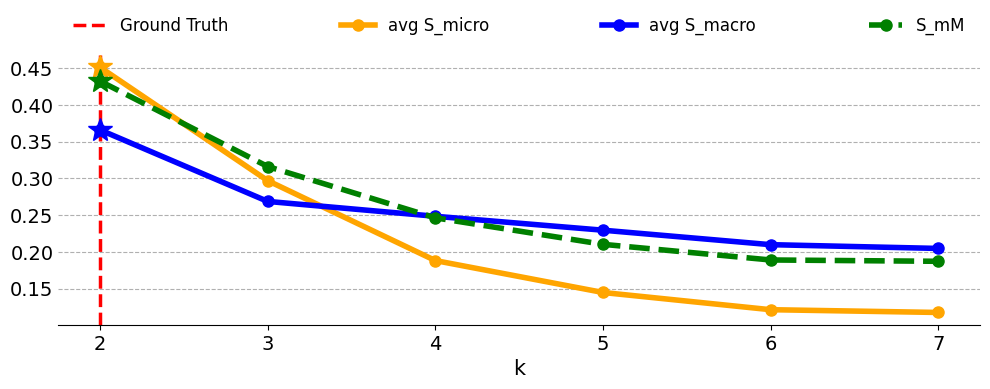}
        \caption{\textsc{Spm}}
        \label{fig:spmtrends}
    \end{subfigure}

    \vspace{0.8em}

    \begin{subfigure}[t]{0.95\textwidth}
        \centering
        \includegraphics[width=\textwidth]{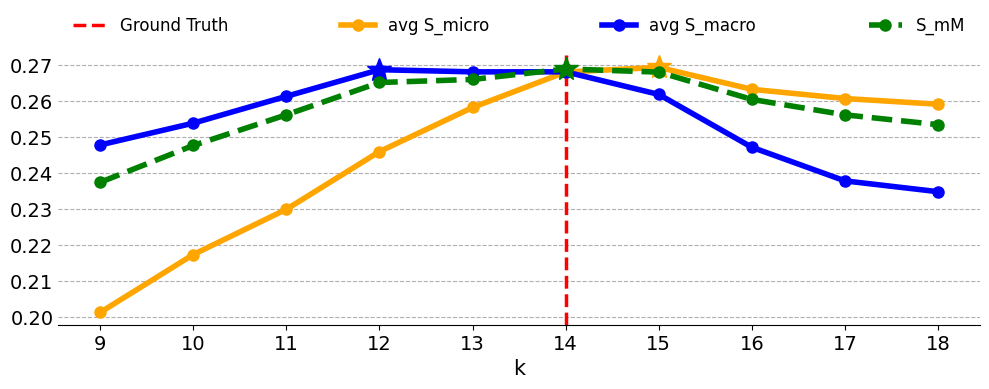}
        \caption{\textsc{Mds}}
        \label{fig:mdstrends}
    \end{subfigure}

    \vspace{0.8em}

    \begin{subfigure}[t]{0.95\textwidth}
        \centering
        \includegraphics[width=\textwidth]{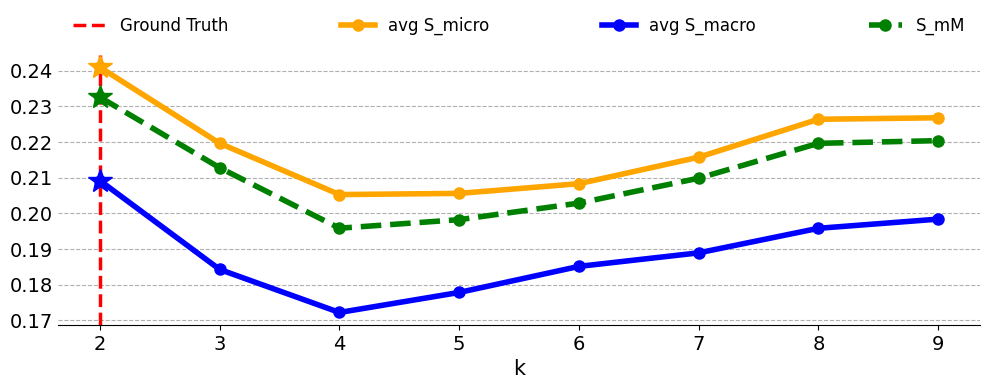}
        \caption{\textsc{Bnk}}
        \label{fig:bnktrends}
    \end{subfigure}

    \caption{Trends of \textcolor{OliveGreen}{$S_\mathrm{mM}$}, \textcolor{orange}{$\overline S_\mathrm{m}$}, and \textcolor{blue}{$\overline S_\mathrm{M}$} over the candidate numbers of clusters $k$. Stars ($\star$) denote the optimal cluster count selected by each criterion. The \textcolor{red}{ dashed line} denotes the ground-truth number of clusters. Datasets: \textsc{Spm}, \textsc{Mds}, and \textsc{Bnk}.}
    \label{app:fig:trends5}
\end{figure}

\begin{figure}[t]
    \centering

    \begin{subfigure}[t]{0.95\textwidth}
        \centering
        \includegraphics[width=\textwidth]{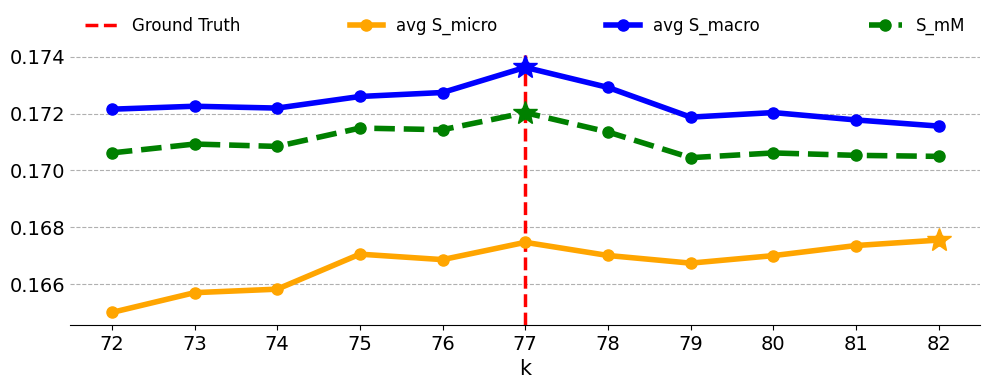}
        \caption{\textsc{B77}}
        \label{fig:b77trends}
    \end{subfigure}

    \caption{Trends of \textcolor{OliveGreen}{$S_\mathrm{mM}$}, \textcolor{orange}{$\overline S_\mathrm{m}$}, and \textcolor{blue}{$\overline S_\mathrm{M}$} over the candidate numbers of clusters $k$. Stars ($\star$) denote the optimal cluster count selected by each criterion. The \textcolor{red}{ dashed line} denotes the ground-truth number of clusters. Dataset: \textsc{B77}.}
    \label{app:fig:trends6}
\end{figure}

\end{appendices}

\end{document}